\documentclass[conference]{IEEEtran}
\IEEEoverridecommandlockouts
\usepackage{cite}
\usepackage{amsmath,amssymb,amsfonts}
\usepackage{algorithmic}
\usepackage[ruled,vlined,linesnumbered]{algorithm2e}
\usepackage{tabularray}
\usepackage{multicol}
\usepackage{multirow}
\usepackage{tabu}
\usepackage{diagbox}
\usepackage{float}
\usepackage{booktabs}
\usepackage{makecell}
\usepackage{subfig}
\usepackage{graphicx}
\usepackage{caption}
\usepackage{textcomp}
\usepackage{hyperref}
\usepackage{url}
\usepackage{xcolor}
\def\BibTeX{{\rm B\kern-.05em{\sc i\kern-.025em b}\kern-.08em
    T\kern-.1667em\lower.7ex\hbox{E}\kern-.125emX}}
\usepackage{bm}

\makeatletter
\renewcommand{\maketag@@@}[1]{\hbox{\m@th\normalsize\normalfont#1}}%
\makeatother
\begin{document}

\title{Graph Condensation for Inductive Node Representation Learning\\
}

\author{
    \IEEEauthorblockN{Xinyi Gao$^{1}$, Tong Chen$^{1}$, Yilong Zang$^{2}$, Wentao Zhang$^{3, 4}$, Quoc Viet Hung Nguyen$^{5}$, Kai Zheng$^{6}$, Hongzhi Yin$^{1*}$}
    \IEEEauthorblockA{$^1$ The University of Queensland, Brisbane, Australia}
    \IEEEauthorblockA{$^2$ German Research Centre for Artificial Intelligence (DFKI), Kaiserslautern, Germany}
     \IEEEauthorblockA{$^3$ Peking University, Beijing, China}
     \IEEEauthorblockA{$^4$ National Engineering Laboratory for Big Data Analysis and Applications, Beijing, China}
    \IEEEauthorblockA{$^5$  Griffith University, Gold Coast, Australia}
    \IEEEauthorblockA{$^6$ University of Electronic Science and Technology of China, Chengdu, China}
    \thanks{* Corresponding author. E-mail addresses: h.yin1@uq.edu.au.}
}

\maketitle

\begin{abstract}

Graph neural networks (GNNs) encounter significant computational challenges when handling large-scale graphs, which severely restricts their efficacy across diverse applications. To address this limitation, graph condensation has emerged as a promising technique, which constructs a small synthetic graph for efficiently training GNNs while retaining performance. However, due to the topology structure among nodes, graph condensation is limited to condensing only the observed training nodes and their corresponding structure, thus lacking the ability to effectively handle the unseen data. Consequently, the original large graph is still required in the inference stage to perform message passing to inductive nodes, resulting in substantial computational demands. To overcome this issue, we propose mapping-aware graph condensation (MCond), explicitly learning the one-to-many node mapping from original nodes to synthetic nodes to seamlessly integrate new nodes into the synthetic graph for inductive representation learning. This enables direct information propagation on the synthetic graph, which is much more efficient than on the original large graph. Specifically, MCond employs an alternating optimization scheme with innovative loss terms from transductive and inductive perspectives, facilitating the mutual promotion between graph condensation and node mapping learning. Extensive experiments demonstrate the efficacy of our approach in inductive inference. On the Reddit dataset, MCond achieves up to $121.5\times$ inference speedup and $55.9\times$ reduction in storage requirements compared with counterparts based on the original graph.

\end{abstract}


\section{Introduction}
Graph neural networks (GNNs) \cite{wu2022graph, zhang2020reliable, zhang2022pasca, zheng2016keyword, sun2021heterogeneous, chandramouli2011streamrec, wu2019session, wang2019semi, 101145, zheng2023automl, yu2023self} have emerged as powerful models for addressing a diverse array of real-world problems. 
Taking node features and their connectivity as the input, GNNs excel at capturing complex dependencies and relationships within graph-structured data, enabling them to extract expressive node representations for downstream tasks~\cite{zhang2021graph, zang2023don, long2023decentralized, nguyen2017argument, long2023model, xia2023efficient, xia2022device, dai2022spatio, xiao2023combining, wang2021fast,li2021lightweight}, e.g., node classification and link prediction. 

However, the increasing prevalence of large-scale graphs poses a significant challenge to GNNs owing to their computational footprints. Most GNNs follow the message passing paradigm \cite{gilmer2017neural} formulated as convolutions over the entire graph, which becomes the major bottleneck \cite{zhang2022graph, xia2023towards} for scaling GNNs to large graphs. Taking the prominent graph convolutional network (GCN) \cite{DBLP:conf/iclr/KipfW17} as an example, its time and space consumptions are respectively quadratic and linear to the number of nodes, which is directly associated with the graph size. This issue is further amplified when multiple versions of one GNN need to be trained, such as neural architecture search~\cite{zhang2022pasca}, continual learning~\cite{rebuffi2017icarl, yang2023time}, and hyper-parameter tuning~\cite{DBLP:conf/kdd/LiSZCJLJG0Y0021} -- all are commonly seen in today's applications.

The efficiency barrier has motivated recent advances in graph condensation (GC)~\cite{jin2022graph,jin2022condensing,loukas2018spectrally}. GC aims to construct a synthetic and compact (e.g., $1,000\times$ smaller) graph that captures essential characteristics of the original full-size graph. In this paper, \textit{original graph} and \textit{synthetic graph} are respectively used to denote the graphs before and after the GC process. The compact synthetic graph allows a GNN to be trained more efficiently, with performance comparable to models trained on the original graph.
Paired with GC, GNNs are endowed with stronger practicality in time-critical and resource-constrained settings. Among the downstream application settings, our main focus in this paper is \textit{inductive node representation learning}~\cite{hamilton2017inductive, DBLP:conf/iclr/ZengZSKP20}, which handles unseen nodes outside the training graph (i.e., the original graph in GC), and is arguably more practical in high throughput systems.

\begin{figure}[t]
\centering
\includegraphics[width=.9\linewidth]{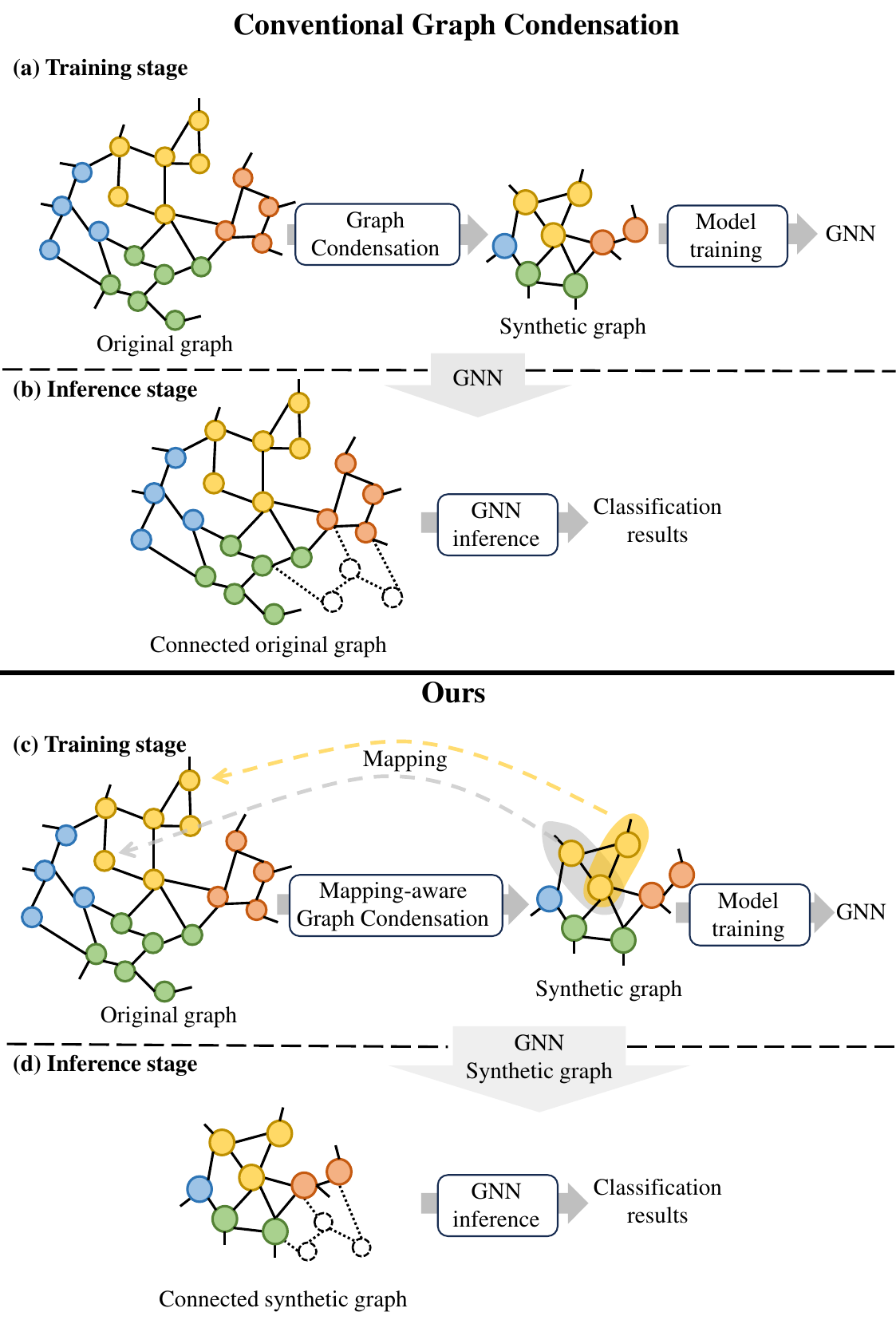}
\caption{The comparison of conventional graph condensation and our proposed method. (a) The training stage of conventional graph condensation. GNN is trained on the small-synthetic graph. (b) The inference stage of conventional graph condensation. GNN performs inference on the original graph with connected test nodes. (c) The training stage of our proposed MCond. Each original node is viewed as a weighted ensemble of some synthetic nodes. (d) The inference stage of MCond. The test nodes are connected on the small-synthetic graph for inference.}
\label{fig:11}
\end{figure}

Despite the potential in promoting scalable use of GNNs, existing GC practices are still suboptimal for inductive inference. Generally, the aforementioned GC methods are centered around GNNs' efficiency during the training stage, while less attention has been paid to the inference/deployment stage, especially in the presence of inductive nodes. In a nutshell, the only objective of their GC process is abstracting node information (i.e., features and labels) and graph structure (i.e., node adjacency) to facilitate cost-effective GNN training, as depicted by Fig. \ref{fig:11} (a). However, to perform inference on inductive nodes, the trained GNN is still required to be deployed on the original graph~\cite{si2022serving,wang2018streaming,chen2013terec}. As shown in Fig. \ref{fig:11} (b), only after connecting with some nodes in the original graph, the message passing to inductive nodes can be performed to learn their representations. 
As a result, inductive nodes cannot be handled without storing the original graph and performing message passing on it, defeating the purpose of abstracting a high-quality, substantially smaller synthetic graph. 

The underlying cause of this defect is that,
the learned synthetic graph is essentially a stand-alone data simulator that resembles the distribution of the original graph w.r.t. node features, structural properties, class labels, etc., but withholds little connection with the nodes in the original graph. Consequently, the synthetic graph cannot establish appropriate and meaningful edges to new nodes, prohibiting inductive message passing purely based on the synthetic graph. 
In an ideal GC process, an explicit mapping between original and synthetic nodes is expected to be learned along with the synthetic graph, then the inductive and synthetic nodes can be easily linked by referring to the original nodes that associate with both sides. 

Hence, to empower GC with stronger inference efficiency for inductive node representation learning, we aim to investigate GC 
with updated objectives to be learned from the original graph: (1) a synthetic graph consisting of simulated nodes, their features, and class labels, as well as the adjacency matrix; (2) a deterministic and symmetric mapping between synthetic and original nodes. The problem studied in this paper is one step above most existing work on GC \cite{jin2022graph,jin2022condensing} which only concerns (1). 
To do so, {the first challenge comes from the parameterization and optimization of the mapping required for (2)}. A recent attempt \cite{si2022serving} is to perform clustering on original nodes, where nodes within each cluster are assigned to the same synthetic node. However, as there are substantially fewer synthetic nodes than original nodes, this will inevitably render multiple inductive nodes having the same set of neighbors in the synthetic graph, thus weakening the uniqueness and expressiveness of their learned representations. { In the meantime, the second challenge involves the joint optimization of these two heavily entangled GC components, as the quality of either the synthetic graph or the mapping bootstraps the other, and will eventually impact the inductive inference performance.}

In light of the challenges, we propose \underline{m}apping-aware graph \underline{cond}ensation (MCond), a novel graph condensation approach with one-to-many node mapping from original nodes to synthetic nodes, to support efficient inductive node representation learning. As Fig. \ref{fig:11} (c) shows, to address the first challenge with MCond, we view each original node as a weighted ensemble of some synthetic nodes, where a sparse $N\times N'$ mapping matrix is learned to encode such information. Here, $N$ and $N'$ are respectively the numbers of original and synthetic nodes ($N'\ll N$). Then, for each inductive node, by looking up its neighbors in the full graph and those neighbors' mapping to synthetic nodes, weighted edges can be established between it and the synthetic graph (Fig. \ref{fig:11} (d)). This brings a significant efficiency boost by involving only the synthetic graph for inductive node inference. 
To tackle the second challenge, we introduce an alternating optimization paradigm to jointly learn the synthetic graph and mapping. Specifically, the synthetic graph is updated by promoting its capability to preserve the structural property and to mirror a GNN's learning trajectory (i.e., gradients) w.r.t. the original graph. Then, by introducing two innovative loss terms from transductive and inductive perspectives, the explicit mapping between nodes can be updated based on the interim synthetic graphs learned from earlier steps. The two optimization procedures are performed alternately by fixing one component and updating the other, so as to collaboratively maximize their synergy and utility.

The main contributions of this paper are three-fold:
\begin{itemize}
  \item We identify the efficiency bottleneck of existing GC methods for inductive node representation learning, and point out the necessity of learning an explicit mapping between original and synthetic nodes, which is an important yet underexplored problem in GC.
  \item We present MCond, a novel GC approach that explicitly learns a sparse mapping matrix to seamlessly integrate new nodes into the synthetic graph for inductive representation learning. In MCond, an alternating optimization scheme is designed to let the synthetic graph and mapping matrix take turns to update toward dedicated objectives. 
  \item Through extensive experimentation, we verify that MCond is both efficient and performant in inductive inference, with up to $121.5\times$ inference speedup and $55.9\times$ reduction in storage requirements compared with counterparts based on the original graph. Additionally, by innovatively introducing two non-parametric calibration methods, label and error propagation, we further validate the effectiveness of the generated synthetic graph and node mapping in capturing valuable structural information.
\end{itemize}

\section{Preliminaries}
In this section, we first revisit both graph neural networks (GNNs) and conventional graph condensation (GC), then present the definition of the problem studied. 

\begin{figure*}[t]
\centering
\includegraphics[width=.9\linewidth]{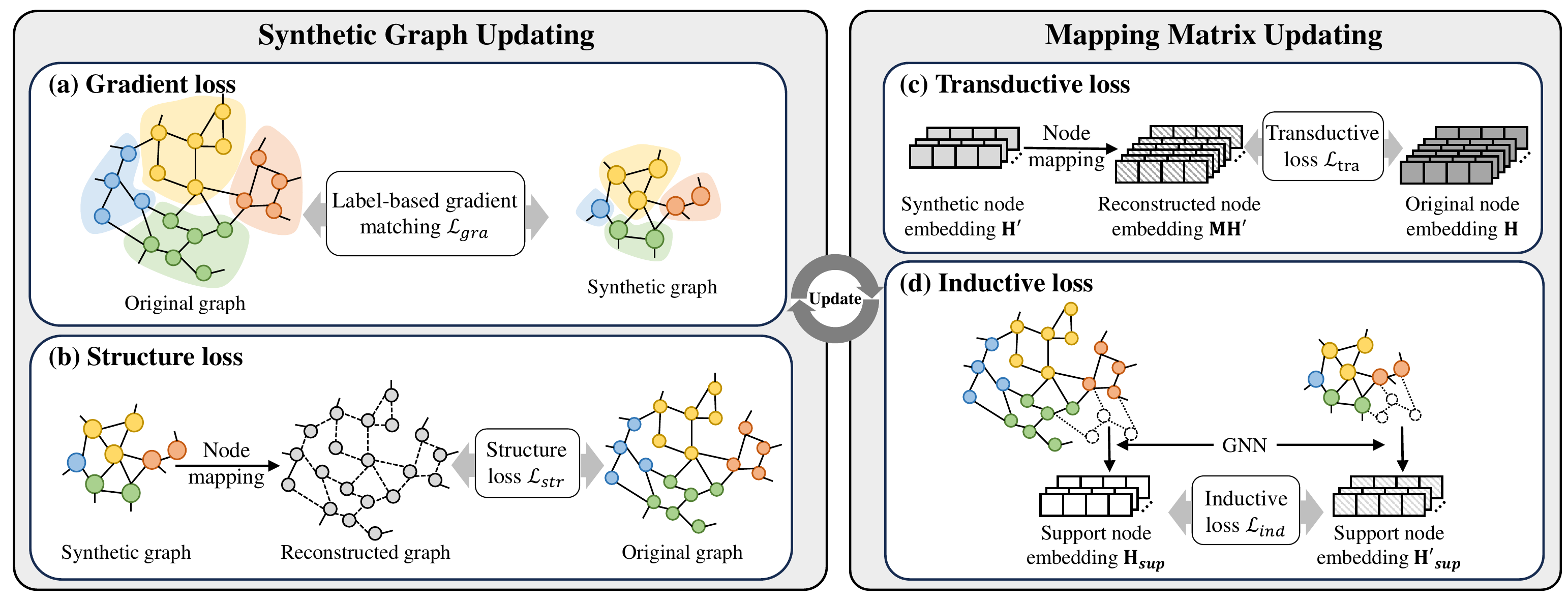}
\caption{The training framework of our proposed MCond. The training procedure involves iterative updates for the synthetic graph and mapping matrix. The synthetic graph is updated using gradient loss and structure loss, which utilize label and structure information as supervision signals, respectively. The mapping matrix is updated using transductive loss and inductive loss, targeting the original node embeddings and connected node embeddings in the original graph, respectively.}
\label{fig:21}
\end{figure*}

\subsection{Graph Neural Networks}
\label{sec:gcn}
We consider a graph $\mathcal{T}=\{{\bf A}, {\bf X}, {\bf Y}\}$ consisting of $N$ nodes, where ${\bf X}\in{\mathbb{R}^{N\times d}}$ is the $d$-dimensional node feature matrix and ${\bf Y}\in\{1,\ldots,C\}^N$ denotes the node labels over $C$ classes. ${\bf A}\in \mathbb{R}^{N\times N}$ is the adjacency matrix, with entry ${\bf A}_{i,j}>0$ denoting an observed edge from node $i$ to $j$, and ${\bf A}_{i,j}=0$ otherwise. 
GNNs leverage the graph topological information and node features to learn node representation (i.e., embeddings) via message-passing. Without loss of generality, we use graph convolutional network (GCN) \cite{DBLP:conf/iclr/KipfW17} as an example, where the propagation process in the $\ell$-th layer is as follows: 
\begin{equation}
{\bf{H}^{(\ell)}} =\text{ReLU}\left(\hat{\mathbf{A}}\bf{H}^{(\ell-1)}\mathbf{W}^{(\ell)}\right),   
\label{eq_GCN}
\end{equation}
where $\hat{\mathbf{A}}=\widetilde{\mathbf{D}}^{-\frac{1}{2}}\widetilde{\mathbf{A}}\widetilde{\mathbf{D}}^{-\frac{1}{2}}$ is the normalized adjacency matrix. {$\widetilde{\mathbf{A}}$ is the adjacency matrix with the self-loop,} $\widetilde{\mathbf{D}}$ is the degree matrix and $\mathbf{W}^{(\ell)}$ is the trainable weights at layer $\ell$. 
${\bf{H}}^{(\ell)}$ is the output node embeddings from the $\ell$-th layer, and ${\bf{H}}^{(0)}=\mathbf{X}$.
By propagating the node features through a total of $L$ GNN layers, each node's final representation can capture information from connected nodes up to $L$-hop away. For simplicity, we denote an $L$-layer GNN as $\mathbf{H}=f(\mathbf{A},\mathbf{X})$, where $\mathbf{H}$ denotes the final node embeddings. 

\subsection{Graph Condensation}
\label{sec:gc}
Graph condensation~\cite{jin2022graph} is proposed to learn a synthetic graph with $N'\ll{N}$ nodes from the original graph $\mathcal{T}$, denoted by $\mathcal{S}=\{{\bf A'}, {\bf X'},{\bf Y'}\}$ with ${\bf A'}\in\mathbb{R}^{N'\times N'}$, ${\bf X'}\in\mathbb{R}^{N'\times d}$, ${\bf Y'}\in\{1,\ldots,C\}^{N'}$, such that a GNN $f(\cdot)$ solely trained on $\mathcal{S}$ can achieve comparable node classification performance to the one trained on the much larger $\mathcal{T}$ \cite{jin2022graph}: 
\begin{equation}\label{eq_GCloss}
\begin{split}
&\min_{\mathcal{S}}\mathcal{L}\big{(}\mathrm{classifier}(f_{{\theta}_{\mathcal{S}}}({\bf A},{\bf X})), {\bf Y}\big{)}\\
&\text{s.t.} \,\, \theta_{\mathcal{S}}= \underset{\theta}{\mathrm{argmin}}\,  \mathcal{L}\big{(}\mathrm{classifier}(f_{\theta}({\bf A'},{\bf X'})), {\bf Y'}\big{)},
\end{split} 
\end{equation}
where $\mathrm{classifier}(\cdot)$ is a readout layer that maps the learned node embeddings into predicted class labels, 
{$f_{\theta}(\cdot)$ denotes the GNN $f(\cdot)$ parameterized by ${\theta}$,}
and $\mathcal{L}(\cdot)$ is the classification loss (i.e., cross entropy in our case). It is worth noting that, node classification is the most common task in GC, while other tasks like node-level regression and outlier detection can be easily applied with availability of labeled data. 

\textbf{Handling Inductive Nodes with $f_{{\theta}_{\mathcal{S}}}(\cdot)$}. As Eq. (\ref{eq_GCloss}) suggests, obtaining a fully trained GNN $f_{{\theta}_{\mathcal{S}}}(\cdot)$ has become much more computationally friendly because of the small-scale synthetic graph $\mathcal{S}$. Let us now investigate the cost of inductive inference, where a batch of $n$ new nodes join $\mathcal{T}$ during test time. Because inductive nodes are unable to establish edges with the synthetic nodes in $\mathcal{S}$, we have to resort to their connectivity with the original graph and aggregate information from nodes within $\mathcal{T}$. We denote their features as ${\bm x}\in{\mathbb{R}^{n\times d}}$, and use an incremental adjacency matrix ${\bm{a}}\in{\mathbb{R}^{n\times N}}$ to encode their connections with nodes in $\mathcal{T}$. To facilitate the forward pass described in Eq. (\ref{eq_GCN}), the adjacency and feature matrices are updated into:
\begin{equation}
\label{connect_test}
\begin{aligned}
\mathbb{A}  ={\begin{bmatrix}
 \mathbf{A} & \!{\bm{a}}^{\top} \\
 {\bm{a}} & \!\!\!\!\widetilde{{\bm{a}}} 
\end{bmatrix}}, \ \ \ 
\mathbb{X}  ={\begin{bmatrix}
 \mathbf{X}  \\
 {\bm{x}}
\end{bmatrix}},
\end{aligned} 
\end{equation}
where $\mathbb{A}\in{\mathbb{R}^{(N+n)\times (N+n)}}$ and $\mathbb{X}\in{\mathbb{R}^{(N+n)\times d}}$. Based on use case and availability, $\widetilde{\bm{a}}\in{\mathbb{R}^{n\times n}}$ can be leveraged to indicate edges among the test nodes, or be zero-padded.
With Eq. (\ref{connect_test}), we are able to perform inductive inference via $\mathbb{H} = f(\mathbb{A},\mathbb{X})$, and retrieve the inductive nodes' embeddings from output $\mathbb{H}$. Assuming a consistent feature dimension $d$ is applied across all $L$ layers, the memory consumption incurred is asymptotically $\mathcal{O}(||\mathbb{A}||_0+(N+n)d)$, which is mainly for storing the non-zero entries in $\mathbb{A}$ (i.e., {norm-0 of $\mathbb{A}$} or the number of edges) and the node features. Meanwhile, each forward pass has a time complexity of $\mathcal{O}(L(N+n)^2d)$. As $n\ll N$ in most real-world applications and can be thus omitted, the space and time complexity for inductive inference is heavily constrained by the size of the original graph, leading to subpar scalability. 

\subsection{Problem Formulation}
Motivated by the deficiency of existing GC in the inductive setting, we now formulate our research problem. For a large original graph $\mathcal{T}=\{{\bf A}, {\bf X}, {\bf Y}\}$, aside from learning a small synthetic graph $\mathcal{S}=\{{\bf A'}, {\bf X'},{\bf Y'}\}$, we also aim to learn a one-to-many mapping from original nodes to synthetic nodes. 
The mapping is parameterized by a non-negative sparse matrix ${\bf M} \in \mathbb{R}^{N\times N'}$. Specifically, each entry ${\bf M}_{i,j}>0$ if the original node $i$ is associated with synthetic node $j$ and ${\bf M}_{i,j}=0$ otherwise, where a higher value indicates a stronger correlation between them. In the next section, we will unfold the design details for learning both $\mathcal{S}$ and ${\bf M}$, and discuss the efficiency advantages of mapping-aware GC in inductive node representation learning.

\section{METHODOLOGIES}
We hereby present our proposed mapping-aware graph condensation (MCond). We begin with the learning of synthetic graph $\mathcal{S}$ (Fig. \ref{fig:21} (a)-(b)), then move onto tasks for updating the mapping ${\bf M}$ (Fig. \ref{fig:21} (c)-(d)). We will wrap up this section with the alternating optimization paradigm for both components, and the deployment of GNNs on $\mathcal{S}$ and ${\bf M}$ for inductive inference.

\subsection{Label-based Gradient Matching (Learning $\mathcal{S}$)}\label{sec:lbgm}
To meet the condensation objective in Eq. \eqref{eq_GCloss}, we adopt the gradient alignment advocated by \cite{jin2022graph,jin2022condensing} as one task for learning $\mathcal{S}$ from $\mathcal{T}$. As a fundamental component in MCond, we introduce it in a relatively brief vein, and our main contribution lies on subsequent improvements over the existing work. In short, when a GNN is being fitted to either the original graph $\mathcal{T}$ or its synthetic version $\mathcal{S}$, its learning trajectory should exhibit similar patterns. 
To promote this, a natural way is to instantiate a relay GNN $f(\cdot)$, and align its gradients w.r.t. both graphs' labels:
\begin{equation}\label{grad_lossGC}
\begin{split}
& \min _{\mathcal{S}} {\mathop{\mathbb{E}}_{\boldsymbol{\theta}_{0}\sim P_{\boldsymbol{\theta}_0}}} \left[\sum_{t=1}^{T} \mathcal{L}_{gra} (\mathcal{G}^{\mathcal{T}}_{\theta_t}, \mathcal{G}^{\mathcal{S}}_{\theta_t} )\right] \\
& \text{s.t.} \,\, {\theta}_{t+1} = \operatorname{optmizer}_{\theta}(\mathcal{L}(\cdot), f_{{\theta}_t}(\cdot), \mathcal{S}),\\
\end{split}
\end{equation}
where $t$ indexes the training step, and $\theta_t$ is the up-to-date GNN parameter at step $t$. $\mathcal{G}^{\mathcal{T}}_{\theta_t}$ and $\mathcal{G}^{\mathcal{S}}_{\theta_t}$ are respectively the gradients of $\theta_t$ w.r.t. graphs $\mathcal{T}$ and $\mathcal{S}$, i.e.,  $\mathcal{G}^{\mathcal{T}}_{\theta_t}=\nabla_{\theta_t}\mathcal{L}(\mathrm{classifier}(f_{\theta_t}(\mathbf{A},\mathbf{X})), \mathbf{Y})$ and $\mathcal{G}^{\mathcal{S}}_{\theta_t}=\nabla_{\theta_t}\mathcal{L}(\mathrm{classifier}(f_{\theta_t}(\mathbf{A}',\mathbf{X}')), \mathbf{Y}')$.
Each training step updates the relay GNN's parameter with a gradient descent-based optimizer $\operatorname{optmizer}_{\theta}(\cdot)$ and \textit{only} $\mathcal{S}$. By taking different parameter initializations drawn from the distribution $P_{\boldsymbol{\theta}_{0}}$, the learned $\mathcal{S}$ can avoid overfitting a specific initialization.  
Suppose the gradient $\mathcal{G}=\{\mathbf{G}^{(\ell)}\}_{\ell=1}^{L}$ in Eq. \eqref{grad_lossGC} entails all $L$ layers' gradient matrices $\mathbf{G}^{(\ell)}$, the gradient distance $\mathcal{L}_{gra}(\cdot)$ is calculated by summing up all layers' pairwise gradient distances: 
\begin{equation}
\mathcal{L}_{gra}(\mathcal{G},\mathcal{G}')=\sum_{\ell=1}^{L}\sum_{i=1}^{D_{\ell}}\left(1-\mathrm{cosine}(\mathbf{G}^{(\ell)}_i, {\mathbf{G}'_i}^{(\ell)})\right),
\end{equation}
where $\mathrm{cosine}(\cdot)$ is the cosine similarity, $\mathbf{G}^{(\ell)}_i$ and $\mathbf{G}'^{(\ell)}_i$ are the $i$-th column vector (out of all $D_{\ell}$) in the gradient matrix $\mathbf{G}^{(\ell)}$ and $\mathbf{G}'^{(\ell)}$ at layer $\ell$, respectively. To facilitate efficient learning of $\mathcal{S}$, a common practice is to reduce the trainable pieces in the synthetic graph $\mathcal{S}=\{\mathbf{A}', \mathbf{X}', \mathbf{Y}'\}$ to only the node features $\mathbf{X}'$. Concretely, the labels $\mathbf{Y}'$ can be predefined to match the distribution of different classes in the original graph, while each entry in $\mathbf{A}'$ is parameterized by the symmetric affinity between the features of two nodes \cite{jin2022graph}:
\begin{equation}\label{eq:adj}
{\bf A}_{i,j}' = \sigma\left(\frac{{\text{MLP}_{\Phi}([{\bf x}'_i; {\bf x}'_j])} + {\text{MLP}_{\Phi}([{\bf x}'_j; {\bf x}'_i])}}{2}\right),
\end{equation}
where $\text{MLP}_{\Phi}(\cdot)$ is a multi-layer perceptron parameterized by~${\Phi}$, and fed with the concatenation of synthetic node features ${\bf x}'_i$ and ${\bf x}'_j$. $\sigma$ is the sigmoid function.

\subsection{Topology-preserving Graph Condensation (Learning $\mathcal{S}$)}
\label{sec:tpgc}
This is where our own design emerges, starting with a more nuanced mechanism for retaining information from the original graph during GC. Note that the remaining components of MCond also need engagement from the relay GNN, for which we keep using notation $f(\cdot)$ throughout the rest of the paper. The condensation process described in Section \ref{sec:lbgm} predominantly preserves the semantic similarity between $\mathcal{S}$ and $\mathcal{T}$, implied by the fact that the synthetic graph structure $\mathbf{A}'$ is purely determined by the pairwise similarity between generated node features. However, the informative structural signals within $\mathcal{T}$ (specifically $\mathbf{A}$) is largely overlooked, making $\mathcal{S}$ a less plausible proxy of $\mathcal{T}$. 

To allow the topological clues within $\mathcal{T}$ to be carried over into $\mathcal{S}$, we put forward a novel structure-based optimization objective in the GC context. Intuitively, we aim to correctly infer the structural information of the original graph from the synthetic one, in which the mapping matrix $\mathbf{M}$ plays a vital role. In $\bf M$, each row $\mathbf{M}_i$ is a $N'$-dimensional vector, where the non-zero entries indicate weighted correlations between original node $i$ and a collection of synthetic nodes. In other words, each original node $i$ can be regarded as a weighted ensemble of synthetic nodes identified by $\mathbf{M}_i$, where the approximate representation of all original nodes $\widetilde{\bf{H}}$ can be obtained via a weighted look-up operation:
\begin{equation}\label{eq_weightedlookup}
    \widetilde{\bf{H}} = {\bf M}{\bf H}', 
\end{equation}
where ${\bf H}'=f_{\theta_t}(\mathbf{A}',\mathbf{X}')$ is the embeddings of all synthetic nodes obtained from the relay GNN.
With that, the topology of $\mathcal{T}$ is introduced as another supervision signal via a link reconstruction task. Given that our original graphs are all unweighted (i.e., $\mathbf{A}_{i,j}=1$ for observed edges and $\mathbf{A}_{i,j}=0$ for unobserved ones) a binary structure loss $\mathcal{L}_{str}$ is formulated in the following logarithm term:
\begin{equation}
\label{str_loss}
\small
\mathcal{L}_{str}(\mathbf{A},{{\widetilde{\bf{H}}}}) =- \frac{1}{ \left | {\mathcal{B}} \right | } \sum_{\left ( i,j \right ) \in {\mathcal{B}}}\mathbf{A}_{i,j}\log \left(\sigma( 
{\widetilde{{\bf{H}}}}_i{\widetilde{{\bf{H}}}}^{{\mathsf{T}}}_j)\right ),
\end{equation}
where ${\mathcal{B}}$ is a mini-batch consisting of both positive and negative edge samples. $\sigma$ is the sigmoid function. Note that the approximate original node representations ${{\widetilde{\bf{H}}}}$ are recomputed at every training step, ensuring consistency with Eq. \eqref{grad_lossGC}. Hence, by incorporating Eq. (\ref{str_loss}) and omitting the initialization and update of the relay GNN in Eq. (\ref{grad_lossGC}), the merged loss term for $\mathcal{S}$ at each training step is: 
\begin{equation}\label{grad_loss}
\begin{split}
& \mathcal{L}_{\mathcal{S}} = \mathcal{L}_{gra}(\mathcal{G}^{\mathcal{T}}_{\theta_t}, \mathcal{G}_{\theta_t}^{\mathcal{S}}) +  \lambda\mathcal{L}_{str}(\mathbf{A},{{{\widetilde{\bf{H}}}}}),
\end{split}
\end{equation}
where $\lambda$ is a balancing hyperparameter to be tuned. Next, we describe how the mapping matrix $\mathbf{M}$ is learned in MCond.

\subsection{Transductive Mapping Constraint (Learning $\mathbf{M}$)}
\label{sec:fpmf}
Based on our earlier discussions, ${\bf M}$ explicitly encodes the node mapping between the original and synthetic graphs, such that each original node can be represented by corresponding synthetic nodes. To efficiently parameterize the mapping matrix ${\bf{M}}$, we enforce a transductive constraint. Specifically, we aim to leverage the representations of existing/transductive nodes in $\mathcal{T}$ to strengthen the mapping to $\mathcal{S}$. Specifically, such a relational mapping should also be captured in the latent embedding space, advocating ${{\bf{H}}} \approx {\bf M}{\bf H}'$. Note that ${\bf{H}}=f(\mathbf{A},\mathbf{X})$ is the exact embeddings of all original nodes produced by the relay GNN on $\mathcal{T}$, instead of the approximate one in Eq. (\ref{eq_weightedlookup}). We also freeze the parameters of the relay GNN while updating $\mathbf{M}$, hence we omit its subscript ${\theta_t}$ when there is no ambiguity.

Though the above constraint comes with a closed-form solution ${\bf{M}}={\bf{H}}{\bf{H}'}^{-1}$, it is inefficient regarding the computations needed. This is because the inverse of non-square matrix $\bf{H}$ cannot be straightforwardly calculated (will take $\mathcal{O}(n^3)$ even if it is square), and a pseudo-inverse using singular value decomposition has to be used as a workaround. As such, we propose to make $\mathbf{M}$ end-to-end trainable via the transductive loss defined below:
\begin{equation}
\mathcal{L}_{tra}(\mathbf{H}, {\mathbf{H}}', {\mathbf{M}})= \frac{1}{N}  || {{\mathbf{H}}} -{\mathbf{M}}{\mathbf{H}}' ||_{2,1}, 
\end{equation}
where $||\cdot||_{2,1}$ is the $L2,1$ matrix norm.

\subsection{Inductive Mapping Constraint (Learning $\mathbf{M}$)}\label{sec:inductive}
Our expectation on the mapping matrix $\mathbf{M}$ is to let it facilitate reasonable connections between inductive nodes and the synthetic graph, such that the message passing can come directly from the synthetic graph $\mathcal{S}$. 
To prepare $\mathbf{M}$ for inductive inference, we sample a small set of nodes from observed nodes, which are excluded from the synthetic graph generation, and term them as support nodes $\mathcal{T}_{sup}$. 
For notation simplicity, we reuse $n$ to denote the number of inductive nodes carried by $\mathcal{T}_{sup}$, and reuse $\bm a$ and $\bm x$ to respectively denote the support nodes' incremental adjacency matrix and features.
With a process similar to Eq. (\ref{connect_test}), the support nodes with incremental adjacency matrix $\bm a$ and feature $\bm x$ are connected to the \textit{synthetic graph} $\mathcal{S}$ instead of the original graph. As such, Eq. (\ref{connect_test}) is updated to the following:
\begin{equation}
\label{ind_matrix}
\begin{aligned}
\mathbb{A}' ={\begin{bmatrix}
 \mathbf{A}' & ({\bm{a}}{\bf M})^{\top} \\
 {{\bm{a}}{\bf M}} & \widetilde{{\bm{a}}} 
\end{bmatrix}},\ \ \ 
\mathbb{X}'  ={\begin{bmatrix}
 \mathbf{X}'  \\
 {\bm{x}}
\end{bmatrix}}.
\end{aligned} 
\end{equation}
In Eq. (\ref{ind_matrix}), ${\bm{a}}{\bf M}$ essentially produces a $n\times N'$ matrix, where the non-zero entries imply edges between inductive and synthetic nodes. 

Then, for the same support nodes and relay GNN $f(\cdot)$, our objective is to minimize the discrepancy between support nodes' embeddings propagated with the original graph $\mathcal{T}$ and those propagated with the synthetic graph $\mathcal{S}$, which are respectively denoted as ${\mathbf{H}_{sup}}$ and ${\mathbf{H}'_{sup}}$ and derived from  $f(\mathbb{A},\mathbb{X})$ and $f(\mathbb{A}',\mathbb{X}')$.  Therefore, the loss function for this inductive task is defined as:
\begin{equation}
\mathcal{L}_{ind}({\mathbf{H}_{sup}},{\mathbf{H}'_{sup}}, \mathbf{M})=\frac{1}{n} || {{\mathbf{H}_{sup}}} -{{\mathbf{H}'_{sup}}} ||_{2,1}. 
\end{equation}
Finally, the mapping matrix is trained by performing optimization under both transductive and inductive constraints:
\begin{equation}\label{eq:L_M}
\mathcal{L}_{\mathbf{M}} = \mathcal{L}_{tra}(\mathbf{H}, {\mathbf{H}}', {\mathbf{M}})+\beta\mathcal{L}_{ind}({\mathbf{H}_{sup}},{\mathbf{H}'_{sup}}, \mathbf{M}),
\end{equation}
with a hyperparameter $\beta$ controlling both sides' contributions.

\begin{algorithm}[t]
\SetAlgoVlined
\small
\textbf{Input:} Original graph
$\mathcal{T}=\{{\bf A}, {\bf X}, {\bf Y}\}$\\
\textbf{Output:} Synthetic graph
$\mathcal{S}=\{{\bf A}', {\bf X}', {\bf Y}'\}$, mapping $\bf M$\\
Initialize $\bf M$, ${\bf X'}$, and ${\bf Y'}$ \\
\For{$k=1,\ldots,K$}  
{
 $\rhd$  Update synthetic graph\\
Initialize $\boldsymbol{\theta}_0\sim P_{\boldsymbol{\theta}_0}$\\
\For{$t=1,\ldots,T$}  
{        
Compute $\mathcal{L}_{\mathcal{S}}$ with Eq. (\ref{grad_loss})\\
${\bf X'} \leftarrow {\bf X'} -\eta_1 \nabla_{\bf X'} \mathcal{L}_{\mathcal{S}}$\\
${\Phi} \leftarrow {\Phi} -\eta_2 \nabla_{\Phi} \mathcal{L}_{\mathcal{S}}$\\
${\theta}_{t+1}\leftarrow \mathrm{optimizer}_{{\theta}}(\mathcal{L}(\cdot),f_{{\theta}_t}(\cdot),\mathcal{S})$  \\
 }
 $\rhd$  Update mapping matrix \\
\For{$t=1,\ldots,T$}{
Compute $\mathcal{L}_{\mathbf{M}}$ with Eq. (\ref{eq:L_M})\\
${\bf{\mathbf M}} \leftarrow {\bf{\mathbf M}} -\eta_3 \nabla_{\bf{\mathbf M}}\mathcal{L}_{\bf M}$\\
}
}
Sparsify ${\bf A'}$ and ${\bf M}$ with Eq. (\ref{eq:threshold})\\
\caption{Alternating Optimization of MCond}
\label{al}
\end{algorithm}

\subsection{Alternating Optimization and Further Details}\label{sec:aofa}
As per our design of MCond, the synthetic graph and mapping matrix are heavily entangled and can thus interfere with each other during training. 
Therefore, we propose to optimize ${\mathcal{S}}$ and ${\bf{M}}$ in an alternating fashion, which is done by updating ${\mathcal{S}}$ or ${\bf{M}}$ (based on $\mathcal{L}_{\mathcal{S}}$ or $\mathcal{L}_{\mathbf{M}}$) for $T$ iterations at a time with the other side fixed. A summary of the training procedure is provided in Algorithm \ref{al}. Notably, after ${\bf{A}}'$ and ${\bf{M}}$ have been learned, an additional sparsification step is performed on both (line 16) before serving the inductive nodes. The rationale is that, by selectively masking out entries with lower values in ${\bf{A}}'$ and ${\bf{M}}$, we can  further minimize the storage overhead with the sparse $N'\times N'$ and $N\times N'$ matrices resulted. The sparsification is facilitated by applying a threshold across each matrix:
\begin{equation}\label{eq:threshold}
{\bf A}_{i,j}'= \begin{cases}
{\bf A}_{i,j}', \text{ if }  {\bf A}_{i,j}' \geq \mu \\
0 \,\,\,\,\,\,\,\,\,\,\text{ otherwise}
\end{cases}\!\!\!\!\!\!,
{\bf M}_{i,j}= \begin{cases}
{\bf M}_{i,j}, \text{ if }  {\bf M}_{i,j} \geq \delta \\
0, \,\,\,\,\,\,\,\,\,\text{ otherwise}
\end{cases}\!\!\!\!\!\!,
\end{equation}
where $\mu$ and $\delta$ are hyperparameters controlling the sparsity of ${\bf{A}}'$ and ${\bf{M}}$, respectively. 
One important note is that, during training, MCond keeps using the dense version of both matrices to ensure efficient and end-to-end gradient-based optimization. {Hyperparameters are subsequently selected based on validation set performance, with a priority given to accuracy.} 
In what follows, we present some further design details and discussions about MCond. 

\textbf{Initialization and Normalization of ${\bf{M}}$.} 
The mapping matrix enables each original node to be represented by a weighted ensemble of synthetic nodes, and our empirical findings (see Section \ref{clu_ini} for details) show that original nodes exhibit stronger correlations with synthetic nodes in the same class. Hence, to speed up convergence, we propose a class-aware initialization strategy for ${\bf{M}}$.
Specifically, for each original node $i$ and its corresponding row $\mathbf{M}_i$ in the matrix, we set $\mathbf{M}_{i,j}$ to a constant (e.g., 1) if synthetic node $j$ has the same class label with it. The class labels for all synthetic nodes are predefined and will stay fixed throughout the training, as described in Section \ref{sec:lbgm}. We set $\mathbf{M}_{i,j} = 0$ if two nodes mismatch in class labels. Leveraging class information for initializing the mapping can effectively  accelerate the learning process. 
At the same time, to guarantee stronger numerical stability for ${\bf{M}}$ and associated computations, we introduce a row-wise normalization step as the following:
\begin{equation}
\label{map}
{\bf{M}}_{i} \leftarrow \mathrm{ReLU}(\frac{\sigma ({{\bf M}}_i)}{\sum_{j=1}^{N'}\sigma ({\bf{M}}_{i,j})} -\epsilon ),
\end{equation}
where a small constant $\epsilon$ is in use to suppress subtle but noisy weights in the normalized matrix. We find in experiments that this normalization secures convergence, and the normalized (but still dense) mapping $\mathbf{M}$ substitutes its original form for computing the forward pass as per Sections \ref{sec:tpgc}, \ref{sec:fpmf}, \ref{sec:inductive} during training.

\textbf{Inductive Inference with $\mathcal{S}$ and $\mathbf{M}$.} In MCond, the inference w.r.t. $n$ inductive nodes leverages an identical process as in Eq. (\ref{ind_matrix}), where the only difference is that the support nodes are replaced by the actual unseen nodes for testing.  
On the one hand, compared with using the original graph for inductive inference, MCond reduces the space complexity from $\mathcal{O}(||\mathbb{A}||_0+(N+n)d)$ to $\mathcal{O}(||\mathbb{A}'||_0+(N'+n)d)$. Considering $N'\ll N$, and $||\mathbb{A}'||_0 \ll ||\mathbb{A}||_0$ even without sparsification of $\mathbf{A}'$, the improvement on memory consumption is substantial. On the other hand, MCond also benefits from a significant complexity drop from $\mathcal{O}(L(N+n)^2d)$ to only $\mathcal{O}(L(N'+n)^2d)$, bringing considerable inference speed up.

\section{Experiments}

We design comprehensive experiments to validate the effectiveness of our proposed MCond and aim to answer the following questions. 
\textbf{Q1}: Compared to other graph reduction methods, can MCond achieve better inductive inference accuracy?
\textbf{Q2}: Can the deployment of the synthetic graph improve the inference time and storage requirement?
\textbf{Q3}: Can the synthetic graph and mapping matrix extract valuable structural information?
\textbf{Q4}: Can the mapping matrix generalize well to different GNN architectures? 
\textbf{Q5}: How do the different components, i.e., optimization constraints, initialization, sparsification, hyper-parameters affect MCond?

\subsection{Experimental Settings}
\label{exp_set}

\noindent\textbf{Datasets}. 
We evaluate our proposed method on three real-world datasets in different characteristics, including a citation network (Pubmed) \cite{DBLP:conf/iclr/KipfW17}, an image network (Flickr) \cite{DBLP:conf/iclr/ZengZSKP20} and a social network (Reddit) \cite{hamilton2017inductive}. In the citation network, papers from different topics are considered as nodes and the edges are citations among the papers. Flickr contains descriptions and properties of images and the node class is the image category. Reddit is a social network where nodes are posts and comments in different topical communities. We follow the public dataset partition and label rates.  The original graph to be condensed only contains the training nodes and their interconnections.
The detailed descriptions of datasets are provided in Table \ref{tab:data}.

\begin{table}[ht]
\caption{The properties of datasets. The training set is utilized as the original graph to be condensed.}
\label{tab:data}
\resizebox{\linewidth}{!}
{\begin{tabular}{l|rrrrr}
\toprule
Dataset    & \#nodes & \#edges & \#feature & \#class & \#training set\\ \midrule
Pubmed     & 19,717   & 44,338      & 500    & 3        & 18,217  \\      
Flickr     & 89,250   & 899,756     & 500    & 7        & 44,625  \\        
Reddit     & 232,965  & 11,606,919  & 602    & 41       & 153,932 \\       
\bottomrule
\end{tabular}}
\end{table}

\noindent\textbf{Baselines}. We compare our proposed methods to conventional graph condensation, four coreset methods, and a virtual graph generation method. For conventional graph condensation, GNN is trained on the synthetic graph, and the inference is performed on the original graph. 
For coreset and virtual graph generation methods, GNN is trained on the original graph and then we use them to generate the synthetic graph for inference. 

The details of each method are as follows:
(1) GCond~\cite{jin2022graph}: the conventional graph condensation method condenses the large graph to a small synthetic graph for efficient GNN training;
(2) Random~\cite{jin2022graph}: randomly select some training nodes for each class and their corresponding subgraphs; 
(3) Degree~\cite{si2022serving}: select the training nodes by sorting the degree of training nodes in each class and picking the nodes with the highest degrees and their interconnections; 
(4) Herding~\cite{welling2009herding}: pick samples that are closest to the cluster center for each class, which is often used in continual learning ~\cite{rebuffi2017icarl,castro2018end}; 
(5) K-Center~\cite{sener2017active,farahani2009facility}: select the center samples to minimize the largest distance between a sample and its nearest center; 
(6) Virtual nodes graph (VNG)~\cite{si2022serving}: constructing the small virtual graph based on the GNN forward pass reconstruct error, which is only used for inference procedures.
We leverage GNN's latent node embeddings for node selection in Herding and K-Center. {The reduced graph size, labels and adjacency matrix format of all baselines are kept the same as our proposed method for a fair comparison}.

\begin{table*}[ht]
\scriptsize
\centering
\caption{The test accuracy (\%) of baselines and MCond under different reduction ratios. The best accuracies are highlighted.}
\setlength\tabcolsep{3.8pt} 
\resizebox{\linewidth}{!}
{
\begin{tabular}{cccclllllllll}
\toprule
                                             &                                                   &                             & O$\to$O                     & \multicolumn{6}{c}{O$\to$S}                                                                                                                                                                            & \multicolumn{2}{c}{S$\to$O}                                                & \multicolumn{1}{c}{S$\to$S}                    \\
\cmidrule(lr){4-4} \cmidrule(lr){5-10} \cmidrule(lr){11-12} \cmidrule(lr){13-13} 
Dataset                                      & Batch                                     & $r$                         & Whole                       & \multicolumn{1}{c}{Random}     & \multicolumn{1}{c}{Degree} & \multicolumn{1}{c}{Herding}    & \multicolumn{1}{c}{K-Center} & \multicolumn{1}{c}{VNG} & \multicolumn{1}{c}{${{\rm{MCond_{OS}}}}$} & \multicolumn{1}{c}{GCond} & \multicolumn{1}{c}{${{\rm{MCond_{SO}}}}$} & \multicolumn{1}{c}{${{\rm{MCond_{SS}}}}$} \\ \midrule
\multicolumn{1}{c|}{\multirow{4}{*}{Pubmed}} & \multicolumn{1}{c|}{\multirow{2}{*}{Graph}} & \multicolumn{1}{c|}{0.16\%} & \multirow{2}{*}{79.03±0.09} & \multicolumn{1}{c}{74.78±0.25} & 74.88±0.14                 & 75.01±0.24                     & 74.58±0.14                   & 75.03±0.05              & \textbf{78.30±0.01}                            & 76.13±0.08                & { 76.70±0.01}                               & 75.97±0.05                                     \\
\multicolumn{1}{c|}{}                        & \multicolumn{1}{c|}{}                             & \multicolumn{1}{c|}{0.32\%} &                             & 74.88±0.16                     & 75.28±0.12                 & 75.08±0.18                     & 74.66±0.15                   & 76.10±0.09              & \textbf{78.40±0.01}                            & 77.01±0.04                & { 77.70±0.01}                               & 76.94±0.01                                     \\ \cmidrule{2-13} 
\multicolumn{1}{c|}{}                        & \multicolumn{1}{c|}{\multirow{2}{*}{Node}}  & \multicolumn{1}{c|}{0.16\%} & \multirow{2}{*}{78.97±0.05} & 75.38±0.13                     & 75.58±0.44                 & 75.28±0.15                     & 75.19±0.24                   & 75.67±0.05              & \textbf{78.20±0.01}                            & 76.14±0.01                & { 76.23±0.05}                               & 75.93±0.24                                     \\
\multicolumn{1}{c|}{}                        & \multicolumn{1}{c|}{}                             & \multicolumn{1}{c|}{0.32\%} &                             & 75.48±0.14                     & 76.08±0.35                 & 75.48±0.21                     & 75.38±0.18                   & 76.40±0.01              & \textbf{78.35±0.05}                            & 76.81±0.04                & { 77.01±0.01}                               & 76.77±0.12                                     \\ \midrule
\multicolumn{1}{c|}{\multirow{4}{*}{Flickr}} & \multicolumn{1}{c|}{\multirow{2}{*}{Graph}} & \multicolumn{1}{c|}{0.10\%} & \multirow{2}{*}{50.88±0.12} & 43.06±0.41                     & 43.50±0.39                 & 43.02±0.40                     & 43.03±0.37                   & 45.53±0.55              & \textbf{47.57±0.07}                            & 47.03±0.14                & { 47.22±0.16}                               & 46.80±0.49                                     \\
\multicolumn{1}{c|}{}                        & \multicolumn{1}{c|}{}                             & \multicolumn{1}{c|}{0.50\%} &                             & 43.16±0.44                     & 44.35±0.32                 & 43.09±0.42                     & 43.08±0.31                   & 46.79±0.44              & \textbf{48.92±0.27}                            & 48.11±0.13                & { 48.70±0.15}                               & 47.96±0.11                                     \\ \cmidrule{2-13} 
\multicolumn{1}{c|}{}                        & \multicolumn{1}{c|}{\multirow{2}{*}{Node}}  & \multicolumn{1}{c|}{0.10\%} & \multirow{2}{*}{49.93±0.06} & 42.49±1.18                     & 44.34±0.94                 & 42.50±1.18                     & 42.49±1.16                   & 45.31±1.14              & \textbf{47.78±0.65}                            & 47.02±0.13                & { 47.73±0.15}                               & 46.65±0.05                                     \\
\multicolumn{1}{c|}{}                        & \multicolumn{1}{c|}{}                             & \multicolumn{1}{c|}{0.50\%} &                             & 42.70±0.88                     & 44.58±0.73                 & 42.67±0.91                     & 42.61±0.89                   & 46.51±1.04              & \textbf{48.34±0.09}                            & 47.95±0.38                & { 48.13±0.17}                               & 47.66±0.61                                     \\ \midrule
\multicolumn{1}{c|}{\multirow{4}{*}{Reddit}} & \multicolumn{1}{c|}{\multirow{2}{*}{Graph}} & \multicolumn{1}{c|}{0.10\%} & \multirow{2}{*}{94.00±0.02} & 55.77±1.76                     & 55.87±1.42                 & 55.86±1.79                     & 55.36±1.78                   & 79.42±1.13              & \textbf{91.53±0.12}                            & 89.63±0.25                & { 90.06±0.23}                               & 88.77±0.44                                     \\
\multicolumn{1}{c|}{}                        & \multicolumn{1}{c|}{}                             & \multicolumn{1}{c|}{0.50\%} &                             & 58.26±1.68                     & 57.30±1.15                 & \multicolumn{1}{c}{58.52±1.77} & 55.42±1.81                   & 83.62±1.51              & \textbf{91.81±0.08}                            & 91.09±0.14                & { 91.33±0.11}                               & 90.66±0.08                                     \\ \cmidrule{2-13} 
\multicolumn{1}{c|}{}                        & \multicolumn{1}{c|}{\multirow{2}{*}{Node}}  & \multicolumn{1}{c|}{0.10\%} & \multirow{2}{*}{92.80±0.07} & 52.82±0.57                     & 53.91±1.11                 & 52.92±0.58                     & 52.38±0.59                   & 79.02±1.55              & \textbf{90.17±0.07}                            & 89.03±0.25                & { 89.52±0.11}                               & 88.14±0.11                                     \\
\multicolumn{1}{c|}{}                        & \multicolumn{1}{c|}{}                             & \multicolumn{1}{c|}{0.50\%} &                             & 55.67±0.55                     & 55.67±1.70                 & 55.80±0.55                     & 52.43±0.56                   & 83.02±1.25              & \textbf{90.59±0.07}                            & 89.23±0.15                & { 89.92±0.07}                               & 88.33±0.17                                     \\ \bottomrule
\end{tabular}
}
\label{tab:41}
\end{table*}

\begin{figure*}[ht]
\centering
\includegraphics[width=\linewidth]{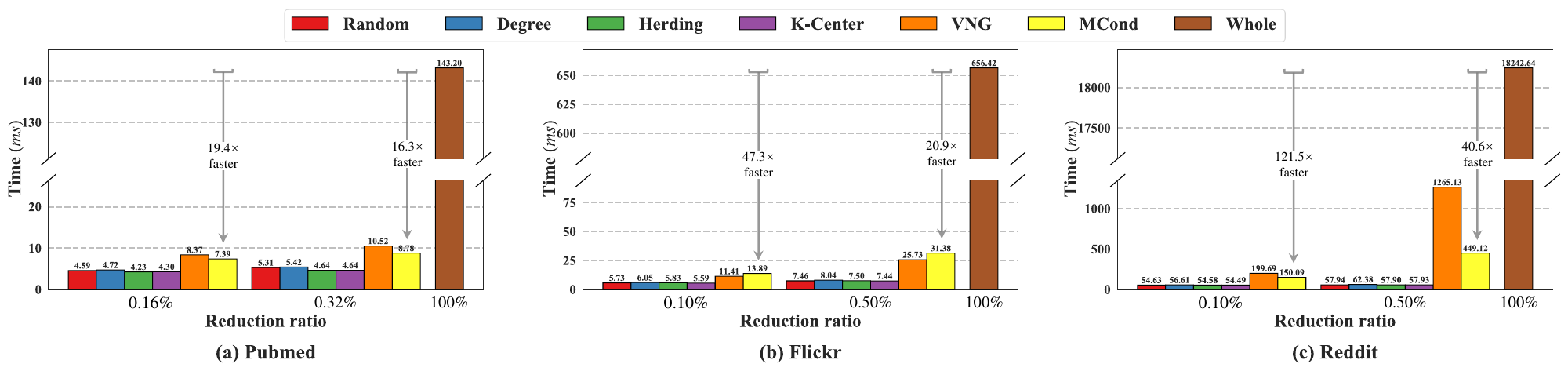}
\includegraphics[width=\linewidth]{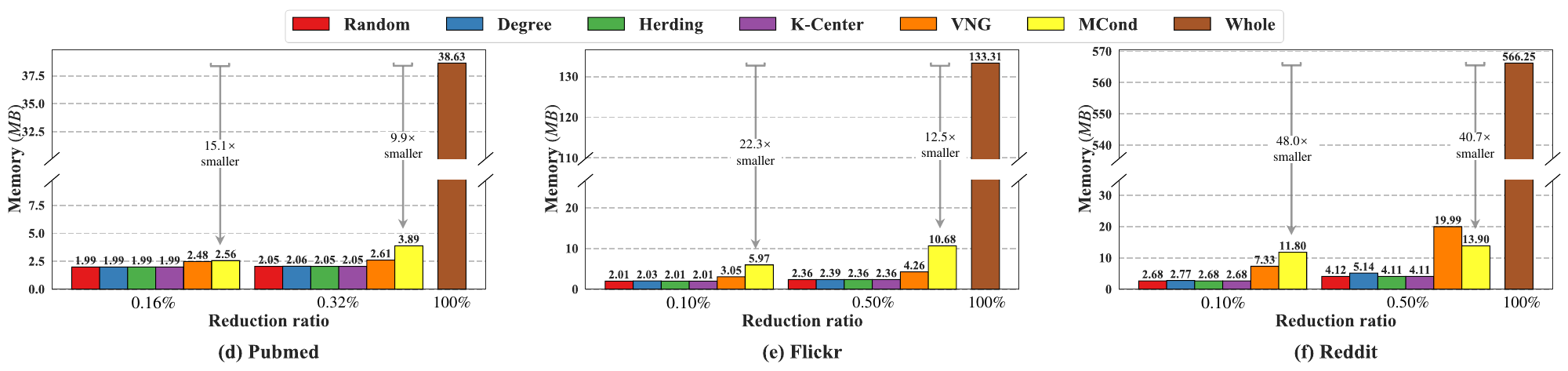}
\caption{Time consumption and memory usage of baselines and MCond for graph batch setting under different $r$. The top three figures represent time consumption and the bottom three figures represent memory usage. We specifically label the acceleration rate and compression rate between MCond and Whole.}
\label{fig:41}
\end{figure*}

\begin{figure*}[t]
\centering
\includegraphics[width=\linewidth]{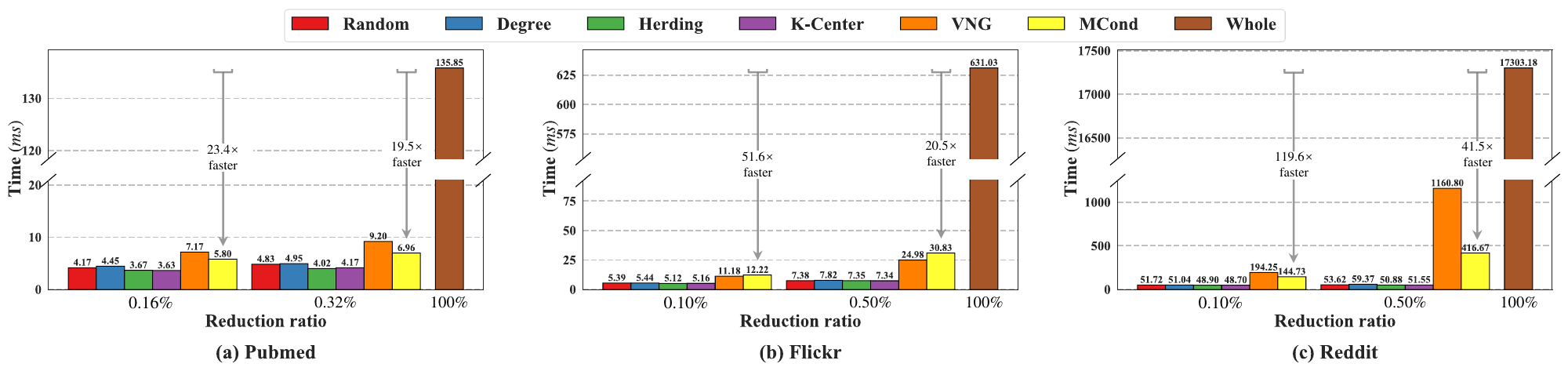}
\includegraphics[width=\linewidth]{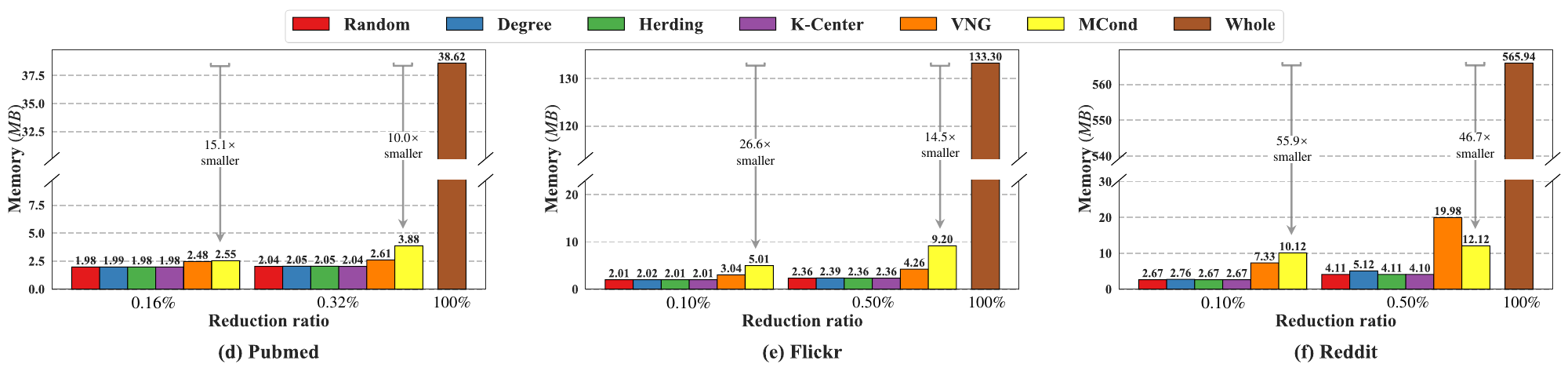}
\caption{Time consumption and memory usage of baselines and MCond for node batch setting under different $r$. The top three figures represent time consumption and the bottom three figures represent memory usage. We specifically label the acceleration rate and compression rate between MCond and Whole.}
\label{fig:411}
\end{figure*}

\noindent\textbf{Evaluation settings}.
\label{eva_set}
As noted in Section \ref{sec:gc}, the $n\times n$ adjacency matrix $\widetilde{\bm a}$ that records the connectivity among the $n$ inductive nodes is optional. Thus, the evaluation of MCond is conducted in two scenarios where inductive nodes come in isolation or a connected subgraph, and we term these settings as \textit{node batch} and \textit{graph batch}, respectively. In the graph batch setting, $\widetilde{\bm a}$ is the adjacency matrix of the inductive nodes.
In the node batch setting, no interconnections between inductive nodes are recorded, hence we zero-out all elements within $\widetilde{\bm{a}}$ for inductive inference. We only have test node features and connections to training nodes.

MCond supports both GNN training and inference on the synthetic graphs. To comprehensively evaluate our approach, we consider four evaluation settings based on the graph deployment: (1) $\text{O}\to \text{O}$ (train and infer on the original graph), (2) $\text{O}\to \text{S}$ (train on the original graph and infer on the synthetic graph), (3) $\text{S}\to \text{O}$ (train on the synthetic graph and infer on the original graph), and (4) $\text{S}\to \text{S}$ (train and infer on the synthetic graph). {In practice, $\text{S}\to \cdot$ is utilized for scenarios with constrained training conditions, such as limited GPU support, restricted memory, and limited training time. $\cdot\to \text{S}$ is useful for latency-sensitive scenarios where fast inference is required.} 
Accordingly, we denote three variants of MCond as follows: (1) $\rm{MCond_{OS}}$: train on the original graph and infer on the synthetic graph, (2) $\rm{MCond_{SO}}$: train on the synthetic graph and infer on the original graph, and (3) $\rm{MCond_{SS}}$: train and infer on the synthetic graph.

We follow the conventional graph condensation \cite{jin2022graph} for the synthetic graph updating setting. Specifically,  SGC~\cite{wu2019simplifying} is adopted for graph condensation and deployment, which utilizes the same graph convolution kernel as GCN \cite{DBLP:conf/iclr/KipfW17} but achieves faster training speed while maintaining comparable performance. 
To evaluate the generalization ability of the synthetic graph and mapping matrix, we evaluate them on various GNN architectures, including SGC~\cite{wu2019simplifying}, GCN~\cite{DBLP:conf/iclr/KipfW17},  GraphSAGE~\cite{hamilton2017inductive}, APPNP~\cite{klicpera2018predict}, and Cheby~\cite{defferrard2016convolutional}.

We condense the original graph with $N$ nodes into a synthetic graph with $N'=r{N}$ ($0<r<1$) nodes, where $r$ is the reduction ratio of synthetic nodes to original nodes.
For the choices of condensation ratio $r$, we choose $r$ of Pubmed to be \{0.16\%, 0.32\%\}, i.e., \{50\%, 100\%\} of its label size considering its sparse label rate. As the nodes in the training set are all labeled in Flickr and Reddit, we choose $r$ as \{0.1\%, 0.5\%\}.

Each method's performance is evaluated using three criteria: test set accuracy (ACC), average inference time per batch (Time), and average memory usage (Memory).

\noindent\textbf{Hyper-parameters and implementation}. 
The hyper-parameters used in synthetic graph generation follow the conventional graph condensation and {others are searched by the grid search method on the validation set to identify the highest validation set accuracy for testing.} 
For all datasets, we use 2-layer models, and the learning rate for updating mapping matrix ${\bf M}$ (line 15 in Algorithm \ref{al}) is 0.1. $\epsilon$ in Eq. \eqref{map} is set as 1e-5. $\alpha$ and $\beta$ are searched from \{0, 0.01, 0.1, 1, 10, 100, 1000\} to balance the losses which differ significantly in magnitude. The number of epochs for Pubmed, Flickr, and Reddit are set as {3000, 4000, 4000}, respectively. 

To follow the common practice of not tuning the model based on the test set, we adopt the validation set as support nodes in the mapping matrix training procedure as \cite{zhao2021adaptive}. It's essential to note that this training process exclusively employs the validation set's features and connections, excluding any labels of the validation set.
We use the ADAM optimization algorithm to train all the models. To eliminate randomness, we repeat each experiment 5 times and report the average test score and standard deviation. The codes are written in Python 3.9 and the operating system is Ubuntu 16.0. We use Pytorch 1.11.0 on CUDA 11.7 to train models on GPU. The inference evaluation is conducted on the CPU with a batch size of 1000. All experiments are conducted on a machine with Intel(R) Xeon(R) CPUs (Gold 5120 @ 2.20GHz) and NVIDIA TITAN RTX GPUs with 24GB GPU memory.

\subsection{Inference Accuracy Comparison (\textbf{Q1})}

We compare MCond with baselines under both node batch setting and graph batch setting, and the average accuracy with standard deviation is shown in Table~\ref{tab:41}. In the $\text{O}\to \text{S}$ setting, $\rm{MCond_{OS}}$ significantly outperforms other baselines. Coreset methods select representative nodes but discard a significant number of edges from the original graph. This action hinders the information propagation in the synthetic graph and impairs inference performance. While VNG generates a virtual graph to replace the original graph for inference, the use of plain weighted k-means and implicit one-to-one mapping relationships in VNG severely limits the expressive capacity of the generated graph, resulting in sub-optimal results. Remarkably, with substantial reduction rates, $\rm{MCond_{OS}}$ achieves results that are closely comparable to Whole, i.e., training and inference on the original graph. Even on the largest dataset Reddit, the accuracy remains well-maintained. Specifically, on the Pubmed dataset, $\rm{MCond_{OS}}$ achieves an accuracy of $78.40\%$ (graph batch) and $78.35\%$ (node batch), closely approaching the accuracy achieved by leveraging the original graph for inference, which is $79.03\%$ and $78.97\%$, respectively.
These results evidently reveal the effectiveness of our proposed mapping matrix, and the synthetic graph can be successfully employed in inductive inference. In the context of the $\text{S}\to \text{O}$ setting, we compare the synthetic graphs generated by $\rm{MCond_{SO}}$ and GCond. The observed improvement in results validates the superior quality of the trained GNN and highlights the effectiveness of incorporating the additional structure loss, which introduces topology information as a valuable supervision signal in the graph condensation process. Lastly, in the comparison between $\rm{MCond_{SO}}$ and $\rm{MCond_{SS}}$, where GNNs are both trained on the synthetic graph, $\rm{MCond_{SS}}$ achieves performances closely similar to $\rm{MCond_{SO}}$. This further indicates that test nodes receive high-quality information from the synthetic graph, and the established edges are appropriate and meaningful for accurate inference.
When comparing the node batch and graph batch settings, the graph batch setting introduces more connections between inductive nodes, resulting in better accuracy across all datasets. {Moreover, ${\bf M}$ ensures that the converted adjacency matrix ${\bm{a}}{\bf M}$ contains more edges than graph batches, exerting dominant influence over information propagation and enabling the node batch setting to maintain high accuracy.}

\subsection{Inference Time and Memory Comparison (\textbf{Q2})}

The inference time and memory usage of Table~\ref{tab:41} are shown in Fig.~\ref{fig:41} and Fig.~\ref{fig:411}.
Due to that $\rm{MCond_{OS}}$ and $\rm{MCond_{SS}}$ both utilize the synthetic graph for inference and yield the same results, we uniformly use ``MCond" to represent the performance on the synthetic graph in Figures. Similarly, ``Whole" represents the inference performance on the original graph, which is the same as GCond and $\rm{MCond_{SO}}$.
We could observe that our proposed method achieves significant inference speedup and memory savings compared to inference on the original graph, particularly on larger graphs and with smaller reduction ratios. On Reddit, MCond achieves $121.5\times$ inference speedup and $48.0\times$ memory saving under the graph batch setting. VNG incurs higher inference time and memory costs on Pubmed and Reddit. The observed differences are due to that VNG generates the dense adjacency matrix, which demands more computations compared to our method. MCond introduces an increase in inference time and memory overhead compared to coreset baselines, primarily due to the utilization of the mapping matrix and the conversion of connections to the synthetic graph.
However, this overhead is efficiently controlled by employing the thresholds to eliminate small entries in the matrix, which enables the sparse matrix operations in the inference procedure and trade-offs between accuracy, storage, and inference speed (refer to Section \ref{abl_stu} for more details). 
Although the graph batch setting achieves better accuracy than the node batch setting, these additional improvements come at the cost of increased memory requirements and higher inference latency. Benefiting from the propagation speedup and graph size reduction, MCond also demonstrates significant improvements in node batch inference. Particularly on the Reddit dataset, the memory savings can be $55.9\times$. This illustrates the generalizability of MCond across different inference settings and makes it a practical solution for inductive inference.

\begin{table}[t]
\scriptsize
\centering
\caption{Test accuracy (\%) and per-time propagation time (ms) of label propagation and error propagation on original (O) and synthetic (S) graphs (acceleration ratio in brackets).}
\setlength\tabcolsep{2pt} 
\resizebox{1\linewidth}{!}{
\begin{tabular}{
ccc cc cc
}
\toprule
\multicolumn{1}{c|}{Dataset ($r$)} & \multicolumn{1}{c|}{Batch} & \multicolumn{1}{c|}{Graph}
 &   Vanilla
 &   LP 
 &   EP
 &   Time   
\\
\midrule
\multicolumn{1}{c|}{\multirow{4}*{\begin{tabular}[c]{@{}c@{}}Pubmed \\ (0.32\%)\end{tabular}}} & \multicolumn{1}{c|}{\multirow{2}*{Graph}} & \multicolumn{1}{c|}{O}   
&    77.70±0.01  &  77.82±0.07  &  77.88±0.10  
&  2.05  
\\
\multicolumn{1}{c|}{}    &\multicolumn{1}{c|}{}    & \multicolumn{1}{c|}{S}
&    76.94±0.01     &  77.92±0.04   &  77.10±0.01   
&  0.54 (3.8$\times$)       
\\

\cmidrule{2-7}
\multicolumn{1}{c|}{}   &\multicolumn{1}{c|}{\multirow{2}*{Node}} & \multicolumn{1}{c|}{O}   
&    77.01±0.01  &  77.34±0.05  &  77.24±0.05  
&  3.75  
\\
\multicolumn{1}{c|}{}    &\multicolumn{1}{c|}{}    & \multicolumn{1}{c|}{S} 
&    76.77±0.12     &  77.84±0.05   &  77.18±0.07   
&  1.39 (2.7$\times$)        
\\

\midrule
\multicolumn{1}{c|}{\multirow{4}*{\begin{tabular}[c]{@{}c@{}}Flickr \\ (0.50\%)\end{tabular}}} &\multicolumn{1}{c|}{\multirow{2}*{Graph}} & \multicolumn{1}{c|}{O}  
&    48.70±0.15  &  49.16±0.33  &  49.07±0.36  
&  2.69  
\\
\multicolumn{1}{c|}{}    &\multicolumn{1}{c|}{}    & \multicolumn{1}{c|}{S} 
&    47.96±0.23     &  48.64±0.24   &  48.43±0.27   
&  0.64 (4.2$\times$)        
\\
\cmidrule{2-7}
\multicolumn{1}{c|}{}    &\multicolumn{1}{c|}{\multirow{2}*{Node}} & \multicolumn{1}{c|}{O}   
&    48.13±0.17  &  48.90±0.12  &  48.50±0.07  
&  5.14  
\\
\multicolumn{1}{c|}{}    &\multicolumn{1}{c|}{}    & \multicolumn{1}{c|}{S}
&    47.66±0.61     &  48.62±0.24   &  48.38±0.26   
&  1.45 (3.5$\times$)        
\\

\midrule
\multicolumn{1}{c|}{\multirow{4}*{\begin{tabular}[c]{@{}c@{}}Reddit \\ (0.10\%)\end{tabular}}} &\multicolumn{1}{c|}{\multirow{2}*{Graph}} & \multicolumn{1}{c|}{O}    
&    90.06±0.11  &  92.74±0.24  &  92.28±0.19  
&  133.43  
\\
\multicolumn{1}{c|}{}    &\multicolumn{1}{c|}{}    & \multicolumn{1}{c|}{S} 
&    88.77±0.44     &  89.82±0.25   &  89.45±0.89   
&  10.21 (13.1$\times$)      
 \\
\cmidrule{2-7}
\multicolumn{1}{c|}{}   &\multicolumn{1}{c|}{\multirow{2}*{Node}}  & \multicolumn{1}{c|}{O}   
&    89.52±0.11  &  91.86±0.03  &  91.48±0.01  
&  130.38  
\\
\multicolumn{1}{c|}{}    &\multicolumn{1}{c|}{}    & \multicolumn{1}{c|}{S}
&    88.14±0.11     &  89.08±0.34   &  88.57±0.37   
&  12.74 (10.2$\times$)      
 \\
 
\bottomrule
\end{tabular}
}
\label{tab:42}
\end{table}

\begin{table*}[ht]
\scriptsize
\centering
\caption{The test accuracy (\%) and inference time (ms) of different GNN architectures.}
\setlength\tabcolsep{4pt} 
\resizebox{0.85\linewidth}{!}
{
\begin{tabular}{
ccc| cc| cc| cc| cc
}
\toprule
\multicolumn{3}{c|}{} 
 &  \multicolumn{2}{c|}{GCN}  
 &  \multicolumn{2}{c|}{GraphSAGE} 
 &  \multicolumn{2}{c|}{APPNP} 
 &  \multicolumn{2}{c}{Cheby}          
\\
\cmidrule{4-11}
Dataset ($r$) & Batch & Method
& Accuracy  & Time 
&  Accuracy  & Time 
& Accuracy  & Time 
&  Accuracy  & Time              
\\
\midrule
\multicolumn{1}{c|}{\multirow{4}*{\begin{tabular}[c]{@{}c@{}}Pubmed \\ (0.32\%)\end{tabular}}} &\multicolumn{1}{c|}{\multirow{2}*{Graph}} & $\rm{MCond_{SO}}$   
&  78.77±0.12  &  79.20  
&  78.40±0.22  &  72.60    
&  77.57±0.12  &  73.94  
&  75.43±0.50  &  81.86 
\\
\multicolumn{1}{c|}{}    &\multicolumn{1}{c|}{}    & $\rm{MCond_{SS}}$
&  76.67±0.54   &  8.72   
&  76.17±0.63   &  9.94    
&  75.03±0.17   &  9.95   
&  73.30±0.36   &  10.28      
\\
\cmidrule{2-11}

\multicolumn{1}{c|}{}&\multicolumn{1}{c|}{\multirow{2}*{Node}}  & $\rm{MCond_{SO}}$   
&  78.70±0.42  &  73.59  
&  79.37±0.31  &  68.61    
&  77.70±1.84  &  66.55  
&  76.97±1.20  &  75.99 
\\
\multicolumn{1}{c|}{}    &\multicolumn{1}{c|}{} & $\rm{MCond_{SS}}$
&  76.37±0.78  &  7.47   
&  78.17±0.12  &  7.09    
&  76.80±1.35  &  6.55   
&  76.37±0.97  &  7.68      
\\

\midrule
\multicolumn{1}{c|}{\multirow{4}*{\begin{tabular}[c]{@{}c@{}}Flickr \\ (0.50\%)\end{tabular}}} &\multicolumn{1}{c|}{\multirow{2}*{Graph}} & $\rm{MCond_{SO}}$    
&  48.54±0.12   &  403.56   
&  48.18±0.29   &  401.67  
&  47.58±0.16   &  401.84  
&  45.54±0.19   &  414.18    
\\
\multicolumn{1}{c|}{}    &\multicolumn{1}{c|}{} & $\rm{MCond_{SS}}$
&  47.38±0.23  &  19.16    
&  47.35±0.20  &  19.26  
&  45.81±0.15  &  19.44  
&  44.45±0.03  &  20.21   
\\

\cmidrule{2-11}
\multicolumn{1}{c|}{} &\multicolumn{1}{c|}{\multirow{2}*{Node}} & $\rm{MCond_{SO}}$   
&  47.52±0.25   &  346.35   
&  47.41±0.17   &  311.50  
&  47.01±0.33   &  358.61  
&  45.22±0.78   &  362.28    
\\
\multicolumn{1}{c|}{}    &\multicolumn{1}{c|}{} & $\rm{MCond_{SS}}$ 
&  47.14±0.43  &  15.73    
&  47.14±0.20  &  19.02  
&  45.54±0.18  &  18.31  
&  44.81±0.77  &  19.10   
\\

\midrule
\multicolumn{1}{c|}{\multirow{4}*{\begin{tabular}[c]{@{}c@{}}Reddit \\ (0.10\%)\end{tabular}}} &\multicolumn{1}{c|}{\multirow{2}*{Graph}} & $\rm{MCond_{SO}}$ 
&  89.26±0.11  &  9719.34  
&  88.89±0.21  &  8657.03     
&  87.12±0.13  &  8504.01  
&  75.30±0.02  &  9860.81  
\\
 \multicolumn{1}{c|}{} & \multicolumn{1}{c|}{} & $\rm{MCond_{SS}}$  
 &  86.73±0.12  &  81.93   
 &  85.82±0.14  &  80.34  
 &  84.26±0.11  &  81.15  
 &  74.12±0.39  &  83.54  
 \\

 \cmidrule{2-11}
 \multicolumn{1}{c|}{} & \multicolumn{1}{c|}{\multirow{2}*{Node}} & $\rm{MCond_{SO}}$ 
&  87.26±0.16  &  9422.38 
&  87.19±0.11  &  8447.82    
&  85.16±0.14  &  8398.37 
&  74.30±0.14  &  9655.12 
\\
 \multicolumn{1}{c|}{} & \multicolumn{1}{c|}{} & $\rm{MCond_{SS}}$ 
 &  84.73±0.14  &  80.30  
 &  83.82±0.16  &  79.41 
 &  83.21±0.13  &  78.92 
 &  72.12±0.16  &  81.94 
 \\
 
\bottomrule
\end{tabular}
}
\label{tab:43}
\end{table*}

\subsection{Label Propagation and Error Propagation (\textbf{Q3})}

Establishing connections between inductive nodes and the small synthetic graph enables the non-parametric calibration of inductive inference on the synthetic graph.
To validate the effectiveness of the generated synthetic graph and node mapping in capturing valuable structural information, we conduct label propagation (LP)~\cite{wang2021combining} and error propagation (EP)~\cite{huang2020combining} techniques on the connected synthetic graph and compare the performance with the original graph results.
Specifically, the LP propagates the synthetic node labels ${\bf Y'}$ to the inductive nodes according to the synthetic graph structure $\mathbf{A}'$ and converted connection, generating predictions for inductive nodes. On the other hand, the EP first calculates the GNN prediction error for synthetic nodes and propagates the errors to revise the GNN predictions of inductive nodes. The small-scale synthetic graph enables high-efficiency propagation. 
We utilize the same GNN model for the original graph and synthetic graph for a fair comparison. The results obtained from the graph batch and node batch setting are presented in Table \ref{tab:42}, and only the propagation time is measured in the evaluation process.
According to the results, the integration of LP and EP consistently leads to improved test accuracies on the synthetic graphs. Notably, the enhanced results even surpass the vanilla performance achieved on the original graph, which becomes even more impressive when considering the negligible propagation time. Furthermore, the graph batch setting demonstrates superior performance compared to the node batch setting, attributed to the additional information obtained from connections among test nodes. These outcomes provide strong evidence that the learned adjacency matrix and converted connections successfully capture essential structural information and can further improve the prediction efficiently, even while retaining the original interconnections between test nodes.

\subsection{Generalizability for Different GNN Architectures (\textbf{Q4})}

An essential property of graph condensation lies in its generalizability across various GNN architectures, allowing the synthetic graph to be employed for training diverse GNN models. To assess the effectiveness of our proposed method in this context, we train different GNN models using the synthetic graph, including GCN, GraphSAGE, APPNP, and Cheby. Subsequently, well-trained GNN modes are leveraged for inductive inference on the connected original and synthetic graph. The test accuracy and inference time under graph batch and node batch settings are presented in Table \ref{tab:43}.
According to the results, all tested GNN models are faster than SGC, as illustrated in Fig. \ref{fig:41} and Fig. \ref{fig:411}. This is due to the fact that these GNN models transform the input features into lower dimensions and then propagate based on the adjacency matrix.
Leveraging the mapping matrix, the test nodes can be successfully inferred by different GNN architectures, yielding comparable inference accuracy and faster inference speed than the results obtained on the original graph. Notably, a performance gap is observed between Cheby and other GNNs on the Reddit dataset, primarily attributed to the limitations of graph condensation in these circumstances. However, even under such circumstances, our proposed method consistently maintains a similar performance to the large graph inference results. This finding demonstrates the effectiveness of the alignment of support node embeddings between the original graph and the synthetic graph, resulting in reliable and comparable inference performance.

\subsection{Ablation Study (\textbf{Q5})}
\label{abl_stu}
We conduct comprehensive experiments to show the effectiveness of each component in our proposed method, including optimization constraints, initialization, sparsification of mapping matrix, and hyper-parameters.

\begin{table*}[ht]
\scriptsize
\centering
\caption{The ablation experiment results of optimization constraints {under ${\rm{MCond_{SS}}}$ setting}.}
\resizebox{0.76\linewidth}{!}{
\begin{tabular}{
c| cc| cc| cc
}
\toprule

 &  \multicolumn{2}{c|}{Pubmed (0.32\%)}  
 &  \multicolumn{2}{c|}{Flickr (0.50\%)} 
 &  \multicolumn{2}{c}{Reddit (0.10\%)}           
\\
\cmidrule{2-7}
Methods 
& Node batch & Graph batch  
& Node batch & Graph batch  
& Node batch & Graph batch    
\\
 \midrule
Plain 
&  74.23±0.17    & 74.35±0.01  
&  46.52±0.68    & 46.82±0.43  
&  79.83±1.29    & 80.52±1.08   
\\
w/o $\mathcal{L}_{str}$
& 76.63±0.05 & 76.53±0.03  
& 47.04±0.38 & 47.79±0.27   
& 84.73±0.40 & 85.48±0.32  
\\
w/o $\mathcal{L}_{ind}$  
& 74.93±0.54  & 76.50±0.08  
& 46.82±0.75  & 47.02±0.24 
& 82.24±1.73  & 82.63±1.03 
\\
${\rm{MCond_{SS}}}$ 
& 76.77±0.12 & 76.94±0.01   
& 47.66±0.61 & 47.96±0.11 
& 88.14±0.11 & 88.77±0.44 
\\
\bottomrule
\end{tabular}
}
\label{tab:44}
\end{table*}

\begin{figure*}[ht]
\centering
\subfloat[]{\includegraphics[width=0.24\linewidth]{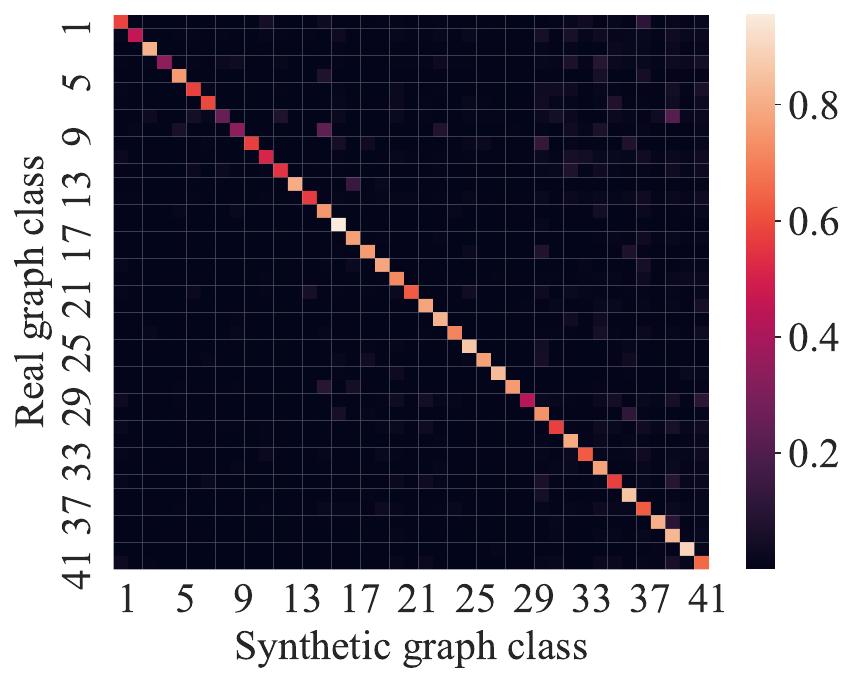} \hspace{0.2cm}}
\subfloat[]{\includegraphics[width=0.24\linewidth]{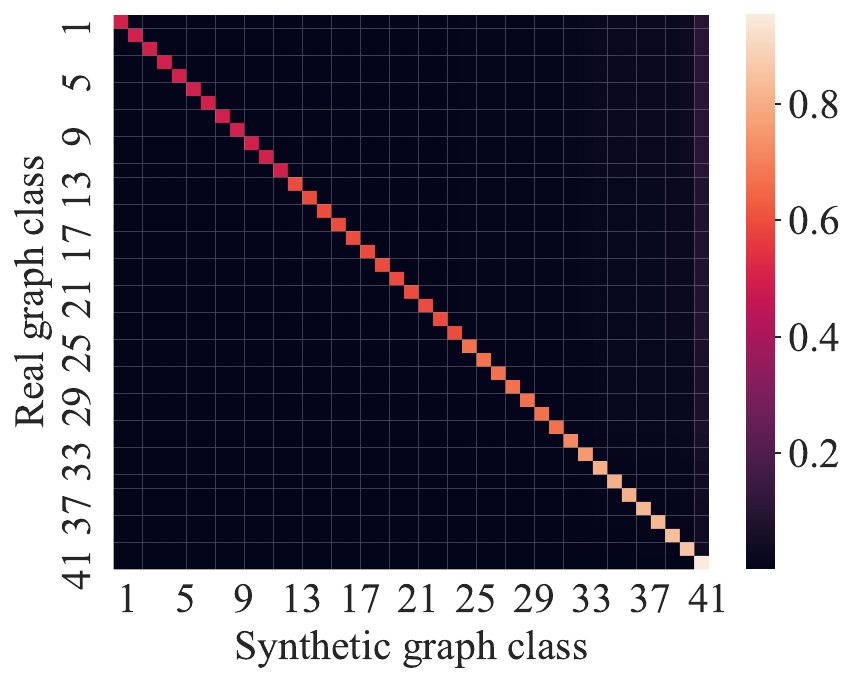} \hspace{0.2cm}}
\subfloat[]{\includegraphics[width=0.3\linewidth]{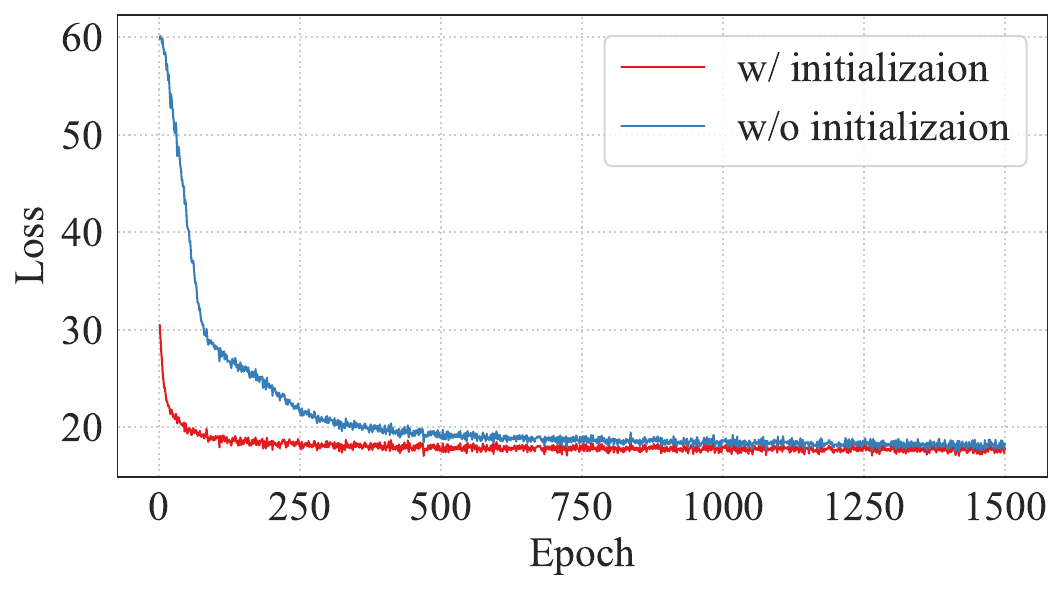}}
\caption{Visualization of mapping matrix $\bf M$ and the effect of initialization on the Reddit dataset {under ${\rm{MCond_{SS}}}$ and node batch setting}. The classes are ranked according to class size. (a) The correlation between original and synthetic graph classes in trained mapping matrix. (b) The correlation between original and synthetic graph classes in the initialized mapping matrix. (c) The comparison of the training loss of the mapping matrix between our proposed initialization and random initialization.}
\label{fig:initial}
\end{figure*}

\noindent\textbf{Optimization constraints.}
To assess the effectiveness of each constraint, ${{\rm{MCond_{SS}}}}$ is evaluated under both graph and node batch settings with the constraints disabled. With $\mathcal{L}_{gra}$ and $\mathcal{L}_{tra}$ serving essential components in generating the synthetic graph and node mapping, we evaluate ${{\rm{MCond_{SS}}}}$ in the following configurations: (i) without both the structure loss $\mathcal{L}_{str}$ and the inductive loss $\mathcal{L}_{ind}$ (referred to as ``Plain"); (ii) without the structure loss $\mathcal{L}_{str}$ (referred to as ``w/o $\mathcal{L}_{str}$"); (iii) without the inductive loss $\mathcal{L}_{ind}$ (referred to as ``w/o $\mathcal{L}_{ind}$"). The results of these settings are presented in Table \ref{tab:44}.
Firstly, when neither the structure loss nor the inductive loss is utilized, the accuracy significantly drops, indicating that both the synthetic graph and the mapping matrix are sub-optimized. Notably, inductive loss emerges as the most influential factor, as it facilitates the simulation of connections of inductive nodes and the alignment of propagated embeddings with the real deployment scenario. This insight emphasizes the crucial role of leveraging the inductive loss in achieving better performance in our proposed method.

\noindent\textbf{Mapping visualization and initialization.}
\label{clu_ini}
We demonstrate the effectiveness of initialization by visualizing the mapping matrix $\bf M$ and the training loss. Fig. \ref{fig:initial} (a) presents the well-trained mapping matrix $\bf M$ of Reddit. It is organized based on the classes to demonstrate the correlation between the classes of the original and synthetic graphs, and classes are ordered by class size.
The values on the diagonal are larger than others, indicating strong self-correlation among classes, and most original nodes are represented by synthetic ones within the same class. Additionally, the off-diagonal values reflect varying degrees of correlation between different classes. For instance, class 9 in the original graph exhibits a significant correlation with class 15 in the synthetic graph.

With this observation, we initialize the mapping matrix through synthetic graph labels, and the resulting initialization is shown in Fig. \ref{fig:initial} (b). In this initialization, higher weights are assigned to the diagonal values, and the color change is attributed to the class imbalance distribution. This indicates that nodes within the same class in the synthetic graph are given higher weights, following the target distribution. Additionally, we compare the loss of the mapping matrix with our proposed initialization and random initialization in Fig. \ref{fig:initial} (c). By using our proposed initialization, the loss starts from a smaller value and decreases faster. Moreover, the initialization leads to a higher accuracy of 88.15\% compared to 87.82\% without initialization.

\begin{figure*}[ht]
\centering
\includegraphics[width=0.84\linewidth]{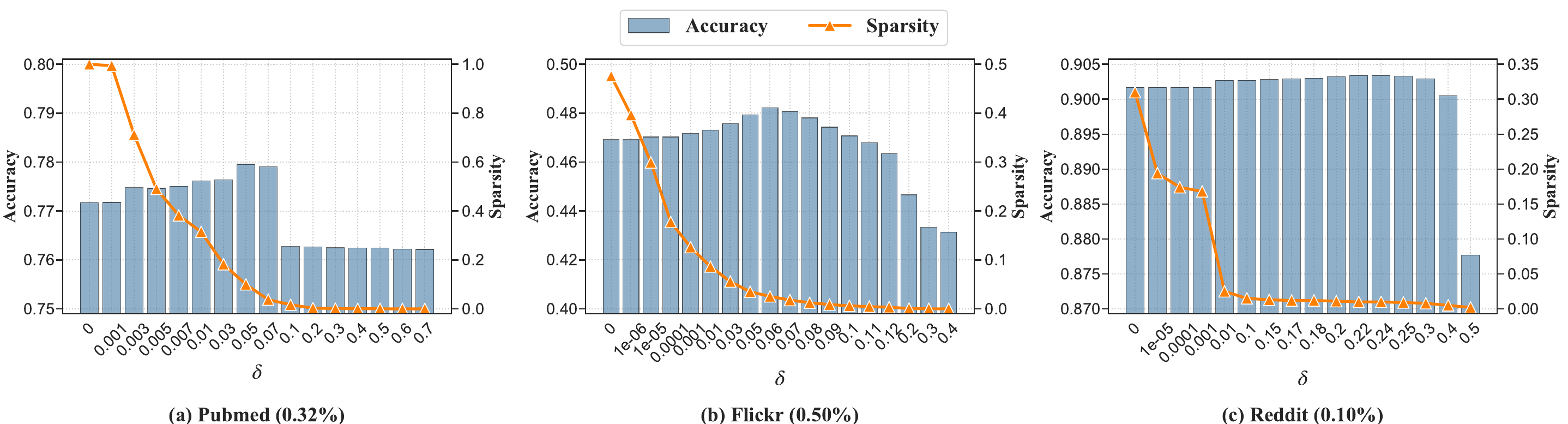}
\caption{Test accuracy and sparsity of mapping matrix under different $\delta$ {for ${\rm{MCond_{OS}}}$ and node batch setting}.}
\label{fig:sparse}
\end{figure*}

\noindent\textbf{Sparsification.}
\label{spar}
To promote the sparsity of the learned mapping matrix $\bf M$, we eliminate entries with values smaller than a given threshold $\delta$. In Fig. \ref{fig:sparse}, we observe the trade-off between sparsity and accuracy with varying $\delta$ on three datasets. As $\delta$ increases, the sparsity of $\bf M$ decreases, while the accuracy initially improves and then declines. This behavior can be attributed to $\delta$ effectively suppressing noise in the mapping matrix and preserving valuable relationships. However, as $\delta$ becomes excessively large, it leads to substantial information loss and results in a decline in performance.

\begin{figure}[ht]
\centering
\includegraphics[width=.8\linewidth]{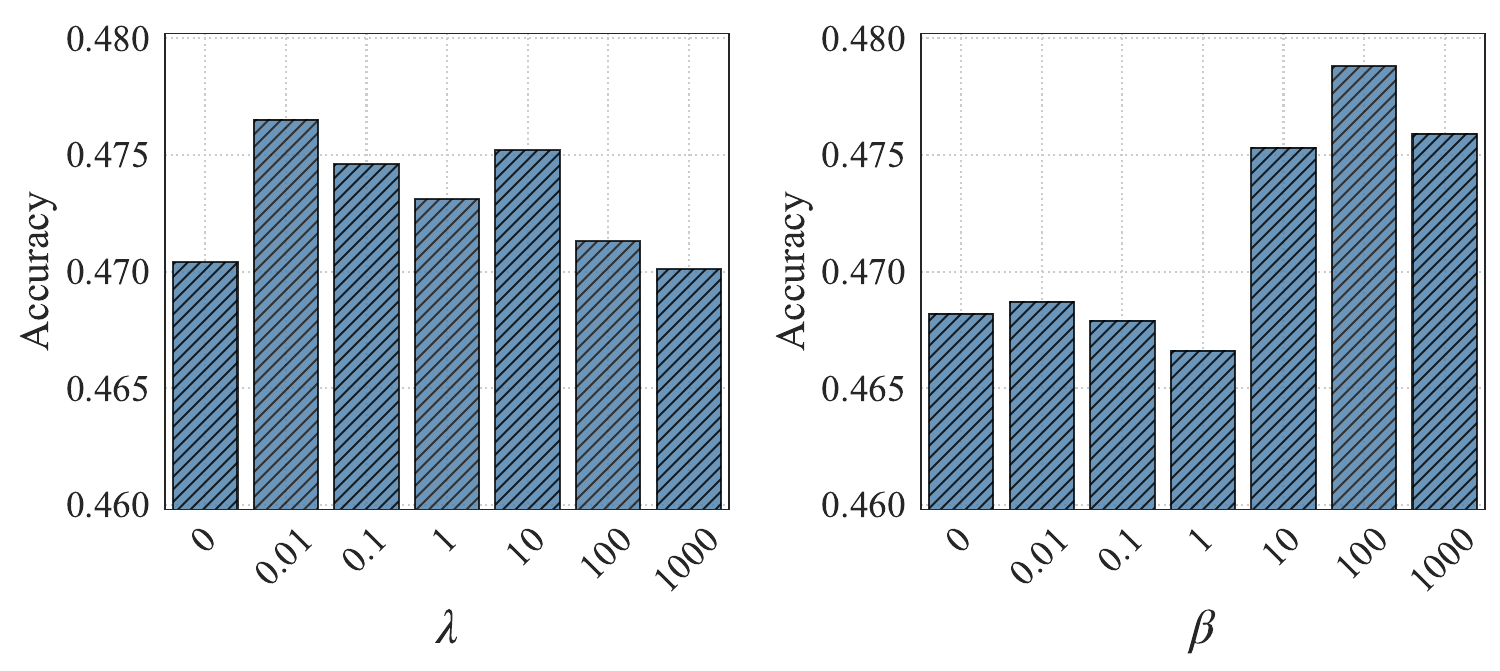}
\caption{The test accuracy of ${\rm{MCond_{OS}}}$ on Flickr ($r$=0.50\%) under different hyper-parameters {for node batch setting}.}
\label{fig:hyper}
\end{figure}

\noindent\textbf{Hyper-parameter sensitivity analysis.}
The hyper-parameters $\lambda$ and $\beta$ play critical roles in controlling the weight of the structure loss and inductive loss during the training procedure. To investigate their influence, we perform the parameter sensitivity experiment on the Flickr dataset. 
In Fig. \ref{fig:hyper}, we present the accuracy performances w.r.t. different values of $\lambda$ and $\beta$. 
Notice that the values of $\lambda$ and $\beta$ vary widely, as different losses and their corresponding gradients differ significantly in magnitude.
For $\lambda$, we observe that values in the range of 0.01 to 0.1 yield the best performance. Higher values may excessively weigh the structure loss, potentially compromising the preservation of label information in the gradient loss.
Regarding $\beta$, a value of 100 ensures a balanced contribution of the inductive loss to the overall training process, resulting in improved performance. 
Overall, to achieve the best performance, $\lambda$ and $\beta$ should be chosen appropriately.

\section{Related Work}

\subsection{Dataset Distillation and Graph Condensation}

The high computation cost of training neural networks on large datasets is a significant concern in the industry. 
In addition to meticulously designed models, researchers have been exploring methods to reduce training costs through data-centric approaches. Dataset distillation (DD)~\cite{wang2018dataset,zhao2020dataset,zhao2023dataset,cazenavette2022dataset} was the first proposed method to distill knowledge from a large training image dataset into a smaller set of synthetic samples. 
As for graph representation learning,  GCond \cite{jin2022graph} introduced the graph condensation method building upon the gradient matching framework. Additionally, DosCond \cite{jin2022condensing} focuses on the challenge of learning discrete structures and fast optimization, particularly in graph classification. However, these methods can only condense the training data and are unable to handle inductive nodes.

\subsection{Coreset Selection and Coarsening}
To reduce the size of training data, it is intuitive to explore the application of coreset selection methods~\cite{har2004coresets}. For example,  
mixture models~\cite{lucic2017training} aim to use Gaussian mixture models to represent the underlying data distribution.
Low-rank approximation~ \cite{cohen2017input} targets finding a low-rank representation of the original dataset that captures its essential characteristics. However, coreset methods may suffer from performance degradation under an extremely small reduction rate.
In contrast, coarsening methods focus on reducing the size of a graph by grouping original nodes into super-nodes and defining their connections. 
Loukas et al. \cite{loukas2018spectrally} provide conditions such that the principal eigenvalues and eigenspaces of coarsened and original graph Laplacian matrices are close. 
Bravo-Hermsdorff et al. \cite{bravo2019unifying} propose a method that simultaneously sparsifies and coarsens graphs, preserving their large-scale structure and Laplacian pseudoinverse. However, coarsening methods mainly focus on reducing the size of a graph without significantly altering its basic properties and do not consider the model and task-specific factors. In contrast, our proposed method is built upon the graph condensation framework and the synthetic graph generated by graph condensation enables training diverse GNN models and achieves comparable performance to models trained on the large graph.

\subsection{GNN inference acceleration}
{Conventional inference acceleration methods typically fall into categories: pruning, quantization, and knowledge distillation (KD). Notably, the advanced pruning methods~\cite{chen2021unified, hui2023rethinking}, simplify both the input graph and GNN weights using learned masks for the adjacency matrix and GNNs. Quantization~\cite{tailor2020degree} employs low-precision integer arithmetic during inference to enhance computational speed. KD techniques~\cite{zhang2021graphless, tian2022nosmog, yang2023learning} aim to train lightweight MLPs that match the performance of the teacher GNN model, resulting in significant inference speedup. However, these methods primarily address model-centric optimizations, while our proposed data-centric approach is orthogonal to them and exhibits better generalization capabilities and broader applicability across various scenarios.}
\section{Conclusion}
In this paper, we present mapping-aware graph condensation (MCond), a novel graph condensation approach that expands the capabilities of the synthetic graph beyond the preservation of GNN training. By introducing a mapping matrix, MCond empowers the synthetic graph seamlessly integrate new nodes into the synthetic graph for inductive representation learning. Extensive experimentation demonstrates the efficacy of MCond in facilitating fast inductive inference across various GNN architectures. 
Moreover, the generated synthetic graph and mapping matrix successfully capture valuable structural information, resulting in further performance enhancements through efficient label propagation and error propagation.
Although the flexibility and compatibility of MCond enable its integration with the graph condensation method, it is important to acknowledge that the performance of our method is significantly influenced by the quality of the graph condensation results. Thus, as a direction for future work, it would be valuable to explore strategies that enhance the robustness of graph condensation for variations in large-scale datasets.
\section*{Acknowledgment}
This work is supported by the Australian Research Council under the streams of Future Fellowship (No. FT210100624) and Discovery Project (DP240101108), as well as the high-performance computing platform of Peking University.
\bibliographystyle{IEEEtran}
\bibliography{IEEEabrv,ref}

\end{document}